\documentclass{article}

\PassOptionsToPackage{numbers, sort&compress, comma, square}{natbib}
    \usepackage[preprint]{neurips_2023}

\usepackage[utf8]{inputenc} 
\usepackage[T1]{fontenc}    
\usepackage{hyperref}       
\usepackage{url}            
\usepackage{booktabs}       
\usepackage{amsfonts}       
\usepackage{nicefrac}       
\usepackage{microtype}      
\usepackage{xcolor}         
\usepackage{graphicx}
\usepackage{multirow}

\title{Language Versatilists vs. Specialists: An Empirical Revisiting on Multilingual Transfer Ability}

\author{
Jiacheng Ye, 
Xijia Tao, 
Lingpeng Kong
\\
The University of Hong Kong \\
\texttt{\{jcye2,xjtao,lpk\}@cs.hku.hk}
}

\begin{document}

\maketitle

\begin{abstract}
Multilingual transfer ability, which reflects how well the models fine-tuned on one source language can be applied to other languages, has been well studied in multilingual pre-trained models (e.g., BLOOM~\citep{scao2022bloom}).
However, such ability has not been investigated for English-centric models (e.g., LLaMA~\citep{touvron2023llama}).
To fill this gap, we study the following research questions.
First, does multilingual transfer ability exist in English-centric models and how does it compare with multilingual pretrained models? Second, does it only appears when English is the source language for the English-centric model? Third, how does it vary in different tasks? 
We take multilingual reasoning ability as our focus and conduct extensive experiments across four types of reasoning tasks. We find that 
the multilingual pretrained model does not always outperform an English-centric model.
Furthermore, English appears to be a less suitable source language, and the choice of source language becomes less important when the English-centric model scales up.
In addition, different types of tasks exhibit different multilingual transfer abilities.
These findings demonstrate that English-centric models not only possess multilingual transfer ability but may even surpass the transferability of multilingual pretrained models if well-trained. By showing the strength and weaknesses, the experiments also provide valuable insights into enhancing multilingual reasoning abilities for the English-centric models.
\end{abstract}
\section{Introduction}
\label{sec:intro}
Multilingual pre-training has become a standard technique to harness the cross-lingual transfer ability of a language model, through which it is possible to improve the performance on low-resource languages by leveraging high-resource languages~\citep{devlin2019bert,conneau2018,conneau2020,lin2022e,scao2022bloom}.
However, there have been looming concerns regarding multilingual pre-training.
For instance, \citet{conneau2020} uncovered \emph{the curse of multilinguality}, suggesting for a fixed model size, cross-lingual performance increases with additional pretraining languages only up to a certain point, after which the performance begins to decline. Additionally, \citet{wang-etal-2020-negative} also reported a phenomenon called \emph{negative interference}, meaning the performance on both high-resource and low-resource languages degrade due to joint multilingual learning.

English-centric models~\citep{brown2020language,chowdhery2022palm,gpt-neo,gpt-j,gpt-neox-20b,biderman2023pythia,zhang2022opt,touvron2023llama}, on the other hand, have demonstrated strong performance on downstream English tasks, but their cross-lingual abilities have not been systematically analyzed.\footnote{In this paper, we refer to a model pre-trained primarily on English corpus as English-centric model.} While it may seem intuitive to assume that English-centric models are not well-suited in cross-lingual transfer, this is not necessarily the case in practice. Research evidence suggests that monolingual models are capable of learning certain abstractions that can generalize across languages, as demonstrated by~\citet{artetxe2020}. In addition, it should be noted that English-centric models are not limited to English only, as they have been exposed to some other languages, albeit to a much lesser extent~\citep{brown2020language,gao2020pile,chowdhery2022palm,touvron2023llama}.

The investigation of multilingual models and English-centric models is especially meaningful in many practical settings. 
Suppose the goal is to develop a model with excellent multilingual reasoning skills such as arithmetic, commonsense, and logical reasoning. In that case, how should we approach this goal? Should we start from an English-centric model which has potentially superior English reasoning abilities and hope these can be transferred to other languages? Or should we start with the multilingual models which generally assumed to have better multilingual transferability, but may lag behind in English reasoning skills?

In this paper, we investigate the following three research questions:
\begin{itemize} 
    \item How does the backbone (e.g., a multilingual pre-trained model or an English-centric model) affect multilingual reasoning?
    \item How does the source language used for downstream task finetuning affect multilingual reasoning on other target languages? For example, will English always be the most effective source language for English-centric models?
    \item How does task type affect multilingual reasoning, e.g., will the reasoning ability be transferred better across languages in some reasoning tasks?
\end{itemize}

To answer these questions, we consider four tasks that require distinct types of reasoning, namely Natural Language Inference, Logical Reasoning, Commonsense Reasoning, and Arithmetic Reasoning, and three popular multilingual and English-centric models, i.e., BLOOM~\citep{scao2022bloom}, Pythia~\citep{biderman2023pythia} and LLaMA~\citep{touvron2023llama}. 
We conduct extensive experiments in these multilingual downstream tasks, and have the following key observations: 
\begin{itemize}
    \item The multilingual pre-trained model does not always outperform an English-centric model, especially for languages seen or rarely seen for both models. For instance, LLaMA achieves a maximum of 9.9\% and a minimum of 0.54\% more average accuracy gain than BLOOM on Turkish and Greek, respectively, both are rarely seen for the two models (\S\ref{sec:rq1});
    \item Incorporating a small amount of multilingual data during the pre-training stage can have a significant impact on English-centric models. For example, LLaMA trained on French and Spanish data with a size of approximately 50 times less than BLOOM, but outperforms BLOOM by up to 23\% on these languages (\S\ref{sec:rq1}); 
    \item The choice of language utilized during fine-tuning becomes less important when the English-centric model scales up (\S\ref{sec:rq2});
    \item Different types of tasks show different multilingual transfer abilities, e.g., logical reasoning knowledge can be transferred better across languages than others. However, as the model size increases, this gap tends to narrow (\S\ref{sec:rq3}).
\end{itemize}

The above findings offers new practical insights into both pre-training and fine-tuning stage of large language models. 
The experiment code is publicly available to promote reproducibility and facilitate further research.\footnote{\href{https://github.com/HKUNLP/multilingual-transfer}{https://github.com/HKUNLP/multilingual-transfer}}


\section{Language Versatilists and Specialists}
In this section, we provide a brief overview of \textbf{1)} multilingual pretraining, with a focus on \textbf{2)} the curse of multilingual pretraining, and then \textbf{3)} discuss the English-centric pretraining, with \textbf{4)} a series of evidence to show the potential of English-centric model possessing multilingual transfer ability. Based on the information presented, we will conclude this section by posing three research questions for investigation.

\paragraph{Multilingual pre-training}
Multilingual pre-training offers a straightforward way to create language versatilists~\citep{devlin2019bert,conneau2018,xue2021,shliazhko2022,lin2022e,scao2022bloom}. The main idea is to combine monolingual corpora in different languages, upsampling those with less data, and training a regular language model on the combined data. After learning multiple languages that use diverse scripts and belong to various language families, the models are expected to possess \textit{cross-lingual transfer ability}, i.e., the model can generalize to target languages~\citep{pires2019,wu-dredze-2019-beto,hu2020xtreme,zhu2023multilingual} when downstream labeled training data is only available in the source language, which is especially important for low-resource target languages~\citep{conneau2018}.

\paragraph{Curse of multilingual pre-training}
\citet{conneau2018} demonstrated that including more languages in a single model can improve performance for low-resource languages but hurt performance for high-resource languages. Furthermore, \citet{wang-etal-2020-negative} shows that negative interference between languages also leads to degraded performance on low-resource languages. As such, prior work had to find a trade-off between supporting more languages and obtaining better performance on a certain set of languages, such as increasing model and vocabulary size~\citep{conneau2018,wang-etal-2020-negative}, and learning additional language-specific parameters through adapters~\citep{pfeiffer2022}.

\begin{table*}[h]
\centering
\caption{Disk size (\textbf{TB}) of the pre-training data per language. 15 languages in the XNLI dataset are shown and sorted by data size in BLOOM.}
\label{tab:pretrain}
\scalebox{1}{
\begin{tabular}{llcc}
\toprule
\textbf{Language} & \textbf{Script} & \textbf{BLOOM} & \textbf{LLaMA} \\
\hline
English (EN) & Latin & 0.485 &  $\sim$4.666 \\
Chinese (ZH) & Chinese ideograms & 0.261 & - \\
French (FR) & Latin & 0.208 & $\sim$0.004 \\
Spanish (ES) & Latin & 0.175 & $\sim$0.004 \\
Arabic (AR) & Arabic & 0.075 & - \\
Vietnamese (VI) & Latin & 0.043 & - \\
Hindi (HI) & Devanagari & 0.025 & - \\
Urdu (UR) & Perso-Arabic & 0.003 & - \\
Swahili (SW) & Latin & \textless{}0.001 & - \\
Bulgarian (BG) & Cyrillic & - & $\sim$0.004 \\
Russian (RU) & Cyrillic & - & $\sim$0.004 \\
German (DE) & Latin & - & $\sim$0.004 \\
Turkish (TR) & Latin & - & - \\
Greek (EL) & Greek & - & - \\
Thai (TH) & Brahmic & - & - \\
\bottomrule
\end{tabular}}
\end{table*}

\paragraph{English-centric pre-training}
While only 13\% of the world’s population speaks
English, the vast majority of NLP research is done
on English. Consequently, numerous models are pre-trained using a corpus that is primarily in English, while without explicitly excluding other languages during data collection~\citep{brown2020language,chowdhery2022palm,gpt-neo,gpt-j,gpt-neox-20b,biderman2023pythia,zhang2022opt,touvron2023llama}. For example, English accounts for approximately 97.4\% in the Pile~\citep{gao2020pile}, an 825GB dataset used by many pre-trained models~\citep{gpt-neo,gpt-j,gpt-neox-20b,biderman2023pythia}, 93\% in training data of GPT-3~\citep{brown2020language}, and around 99\% in training data of LLaMA~\citep{touvron2023llama}. In comparison, the largest constitution, i.e., English, only accounts for 30\% in the ROOTS~\citep{laurenccon2022bigscience}, which is the multilingual corpus for pretraining BLOOM~\citep{scao2022bloom}. Table~\ref{tab:pretrain} compares the data size of the pretraining corpus for BLOOM and LLaMA model across 15 languages from XNLI dataset~\citep{conneau2018xnli}.

\paragraph{Harbingers of multilingual transfer ability in English-centric models}
Multiple lines of evidence suggest that English-centric models have the potential for multilingual transfer capability. 
On the one hand, large English-centric models perform comparably with multilingual models on multilingual question-answering tasks~\citep{chowdhery2022palm} and translating other languages into English~\citep{brown2020language,chowdhery2022palm}, though still lagging behind in translating into other languages. 
On the other hand, prior work suggests that the source of multilingual transfer ability may not be solely attributed to the multilingual pretraining process, as monolingual models also learn some abstractions that generalize across languages~\citep{artetxe2020}.
High-level knowledge-transferring phenomena have been observed in other modalities, such as from English to Python~\citep{hernandez2021scaling}, from `non-linguistic data with grammatical structure' to language~\citep{papadimitriou2020pretraining,ri2022pretraining}, and from language to vision~\citep{lu2021pretrained}. 
Similarly, the presence of innate biological properties of the brain that constrain possible human languages was posited to explain why children learn languages so quickly despite the poverty of the stimulus~\citep{chomsky1981naturalistic,legate2002empirical}.

\paragraph{Research Questions}
After rethinking the challenge of multilingual pretraining and the potential of English-centric training, to what extent multilingual transfer ability exists in English-centric models compared with multilingual pre-trained models remains still unclear. 
In this work, we are particularly interested in the three research questions as listed in \S\ref{sec:intro}.
\section{Experiments}

\subsection{Setup}

\paragraph{Datasets}

We consider the following four types of tasks that require distinct reasoning abilities: 
\begin{itemize}
\item \textbf{Natural Language Inference}: we use XNLI dataset~\citep{conneau2018xnli}, which is created by crowd-translating the dev and test portions of the English Multi-NLI dataset~\citep{williams2018broad} into 14 languages (French (fr), Spanish (ES), German (DE), Greek (EL), Bulgarian (BG), Russian (RU), Turkish (TR), Arabic (AR), Vietnamese (VI), Thai (TH), Chinese (ZH), Hindi (HI), Swahili (SW), and Urdu (UR)); 
\item \textbf{Logical Reasoning}: we adopt LogiQA dataset~\citep{liu2021logiqa}, which is sourced
from expert-written questions for testing human
logical reasoning. As the training set is only available in English and Chinese, we further translate both training and test splits into French with Google Translate API\footnote{\href{https://cloud.google.com/translate}{https://cloud.google.com/translate}};
\item \textbf{Commonsense Reasoning}: we choose XCOPA dataset~\citep{ponti2020xcopa}, which is a causal commonsense reasoning task in which a model is given a premise sentence and must determine either the cause or effect of the premise from two possible choices. Since the dataset only provides multilingual test sets, we utilize the training set from the original English COPA release~\citep{roemmele2011choice} and additionally translate it into Chinese and French with Google Translate API;
\item \textbf{Arithmetic Reasoning}: we use GSM8K dataset~\citep{cobbe2021training}, which contains linguistically diverse grade school math word problems. \citet{shi2022b} construct a multilingual test set which we directly adopt for our test set. To construct a multilingual training set, we further translate the English training set into French and Chinese with Google Translate API. 
\end{itemize}

\paragraph{Models}
We consider both multilingual models and English-centric models and choose the three most popular models as the backbone in our experiments. The details of them are listed as follows:
\begin{itemize}
    \item \textbf{BLOOM}~\citep{scao2022bloom}: a series of models trained on ROOTS~\citep{laurenccon2022bigscience}, a multilingual corpus containing 341 billion tokens from 46 natural languages and 13 programming languages. We consider three model sizes, i.e.,  560M, 1.7B, and 7.1B, in our experiments;
    \item \textbf{Pythia}~\citep{biderman2023pythia}: a family of models trained on the Pile~\citep{gao2020pile}, an English-centric corpus contains 207 billion tokens after deduplication. The overall number of tokens of the deduplicated Pile is on par with ROOTS. We consider three model sizes, i.e., 410M, 1.4B, and 6.5B, in our experiments;
    \item \textbf{LLaMA}~\citep{touvron2023llama}: a series of models trained on various English-centric corpus, summing up to tokens (1.4 trillion), much larger than that in ROOTS (341 billion) and the Pile (207 billion). Currently, LLaMAs are one of the most well-performed open-sourced models among similar-sized models. We consider three model sizes, i.e., 6.7B, 13B, and 32.5B, in our experiments.
\end{itemize}

\paragraph{Implementation Details}
We fine-tune the above 9 models on each of the languages (i.e., 15 languages for XNLI and 3 languages for others). As full fine-tuning becomes less feasible when the model gets larger, we adopt Low-Rank Adaptation (LoRA; ~\citep{hulora}) which freezes the pre-trained model weights and injects trainable rank decomposition matrices into each layer of the Transformer architecture to reduce the number of trainable parameters for downstream tasks.
We add trainable decomposition matrices for the query, key, and value projection matrices in all the self-attention modules. 
As we fix the pre-trained model weights, we further adopt Int8 matrix multiplication~\citep{dettmers2022llmint8} during LoRA finetuning and inference to cut the memory needed.
With the above techniques, the finetuning and inference for the latest 32.5B model can be accomplished on a single NVIDIA A100-80GB GPU.
Additionally, instead of using all the 400k training instances for each language in the XNLI dataset, we limit the number of training instances to 9k, with 3k for each class, to reduce computation.
We set the batch size to 32, the learning rate to 3e-4, and the number of epochs to 3.
We adopt instruction fine-tuning~\citep{weifinetuned,sanhmultitask} instead of classifier-based fine-tuning~\citep{devlin2019bert} for classification tasks, which better injects certain abilities without changing the architecture.
The number of instances and instruction templates for each dataset are listed in the Appendix Table~\ref{tab:prompts}. During inference, we compare the perplexity of each option to decide the label for classification tasks following~\citep{brown2020language,ye2022zerogen}, and we adopt the open-sourced OpenICL toolkit~\citep{wu2023openicl} for implementation.
We always use English prompts as suggested by prior works~\citep{lin2022e,muennighoff2022}.

\subsection{Findings for RQ1}
\label{sec:rq1}

\begin{table}[t]
\scriptsize
\centering
\caption{Accuracy of similar-sized multilingual and English-centric models on each test language after finetuning on English task data. The language is sorted by the pre-train data size in BLOOM as shown in Table~\ref{tab:pretrain}. Ave(15) refers to the average results of all 15 test languages and Ave(3) is the average of the top three resourced languages (EN, ZH, FR) in BLOOM. Best result is in bold for each language. Full results of all model sizes and all the training languages are shown in the Appendix.}
\label{tab:main}
\setlength{\tabcolsep}{0.65mm}
\scalebox{1}{
\begin{tabular}{lccccccccccccccccc}
\toprule
\multicolumn{1}{l}{\multirow{2}{*}{\textbf{Model}}} & \multicolumn{17}{c}{\textbf{XNLI}} \\
\cmidrule(lr){2-18}
\multicolumn{1}{c}{} & \textbf{en} & \textbf{zh} & \textbf{fr} & \textbf{es} & \textbf{ar} & \textbf{vi} & \textbf{hi} & \textbf{ur} & \textbf{sw} & \textbf{bg} & \textbf{ru} & \textbf{de} & \textbf{tr} & \textbf{el} & \textbf{th} & \multicolumn{1}{c}{\textbf{Ave(15)}} & \multicolumn{1}{c}{\textbf{Ave(3)}} \\
\midrule
\textbf{BLOOM-7.1B} & 81.38 & \textbf{70.72} & 75.25 & \textbf{77.96} & \textbf{69.46} & \textbf{69.96} & \textbf{62.75} & \textbf{57.33} & \textbf{56.25} & 50.52 & 59.60 & 59.72 & 44.15 & 51.22 & 46.59 & \textbf{62.19} & \textbf{75.78} \\
\textbf{Pythia-6.9B} & 83.77 & 61.84 & 70.10 & 70.84 & 56.03 & 55.91 & 47.31 & 46.31 & 45.59 & 61.88 & 61.10 & 65.89 & \textbf{54.39} & \textbf{61.50} & \textbf{51.42} & 59.59 & 71.90 \\
\textbf{LLaMA-6.7B} & \textbf{86.85} & 61.82 & \textbf{76.99} & 77.56 & 52.69 & 54.71 & 46.97 & 45.51 & 40.58 & \textbf{72.79} & \textbf{73.09} & \textbf{75.81} & 51.12 & 57.39 & 46.57 & 61.36 & 75.22 \\
\bottomrule
\end{tabular}}

\setlength{\tabcolsep}{2.02mm}
\vspace{2pt}
\scalebox{1}{
\begin{tabular}{lcccccccccccc}
\multicolumn{1}{l}{\multirow{2}{*}{\textbf{Model}}} & \multicolumn{4}{c}{\textbf{GSM8K}} & \multicolumn{4}{c}{\textbf{LogiQA}} & \multicolumn{4}{c}{\textbf{XCOPA}} \\
\cmidrule(lr){2-5}
\cmidrule(lr){6-9}
\cmidrule(lr){10-13}
\multicolumn{1}{c}{} & \textbf{en} & \textbf{zh} & \textbf{fr} & \multicolumn{1}{c}{\textbf{Ave(3)}} & \textbf{en} & \textbf{zh} & \textbf{fr} & \multicolumn{1}{c}{\textbf{Ave(3)}} & \textbf{en} & \textbf{zh} & \textbf{fr} & \multicolumn{1}{c}{\textbf{Ave(3)}} \\
\midrule
\textbf{BLOOM-7.1B} & 11.60 & \textbf{8.80} & 14.00 & 11.47 & 25.81 & 23.35 & 23.96 & 24.37 & 54.00 & 51.40 & 48.80 & 51.40 \\
\textbf{Pythia-6.9B} & 12.80 & 6.00 & 10.00 & 9.60 & 32.10 & 27.96 & 30.26 & 30.11 & 50.80 & 50.00 & 53.40 & 51.40 \\
\textbf{LLaMA-6.7B} & \textbf{27.20} & 7.20 & \textbf{18.00} & \textbf{17.47} & \textbf{37.63} & \textbf{31.34} & \textbf{33.79} & \textbf{34.25} & \textbf{85.60} & \textbf{59.80} & \textbf{71.40} & \textbf{72.27} \\
\bottomrule
\end{tabular}}
\end{table}
\textit{"RQ1: How does the backbone (e.g., a multilingual pre-trained model or an English-centric model) affect multilingual reasoning?"}

\begin{figure}[t]
\centering
\includegraphics[width=5.5in]{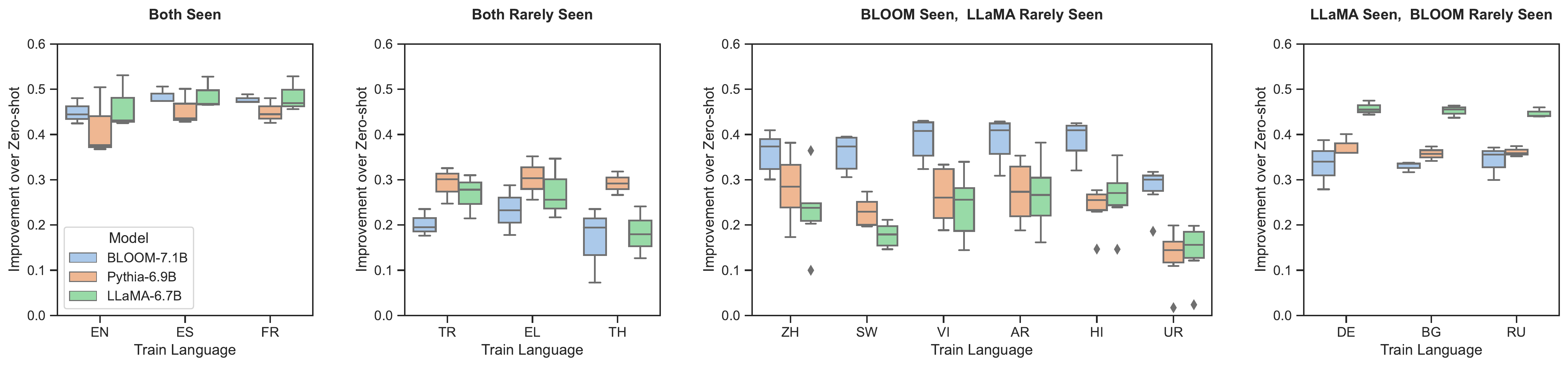}
\caption{Evaluating BLOOM-7.1B and LLaMA-6.7B on four groups of languages, i.e., both seen during pre-training, both rarely seen during pre-training, seen for BLOOM but rarely seen for LLaMA, and seen for LLaMA but rarely seen for BLOOM. The zero-shot accuracy is subtracted to better reflect performance gain brought from additional training on the certain source language.
}
\label{fig:transfer}
\end{figure}

To facilitate the discussion, we use three models of similar parameters, i.e., BLOOM-7.1B, Pythia-6.9B, and LLaMA-6.7B. We begin by showing the overall accuracy of the three models on all the languages after training on English task data, as shown in Table~\ref{tab:main}.
Then, we split the train languages into four categories based on the pretraining languages of BLOOM and LLaMA as listed in Table~\ref{tab:pretrain}: (1) seen for both, (2) rarely seen for both, (3) seen for BLOOM but rarely for LLaMA, and (4) seen for LLaMA but rarely for BLOOM.
We visualize the results in Figure \ref{fig:transfer}, where the zero-shot accuracy is subtracted to better reflect performance gain brought from additional training on the certain source language.

\paragraph{A minimal amount of multilingual data makes a lot in English-centric models}
As shown in Table~\ref{tab:main}, LLaMA achieves comparable or better overall performance on the multilingual test sets, with an average accuracy of 61.39\% compared to 62.19\% of BLOOM. Even on languages frequently seen by BLOOM (i.e., English, Chinese, and French), the average performance of LLaMA can still match (XNLI) or outperform (GSM8K, LogiQA, and XCOPA) BLOOM.
During pre-training, LLaMA only sees French and Spanish data with individual sizes equal to roughly 4 GB. By contrast, BLOOM has seen about 50 times data in these languages in the pre-training stage. Nevertheless, when evaluating LLaMA on French, the accuracy exceeds that of BLOOM by more than 1.7\%, 4\%, 10\%, and 23\% on XNLI, GSM8K, LogiQA and XCOPA, respectively. LLaMA also achieves a very similar accuracy on Spanish, with BLOOM performing slightly better by a margin of 0.4\% on XNLI. 

However, for languages without any pre-training data (e.g., Chinese, Arabic, Vietnamese, etc.), the performance lags behind BLOOM by around 15\% on XNLI but is still comparable or better on other tasks.
Our findings suggest that incorporating a minimal amount of diverse low-resource language data during pre-training can result in a more capable multilingual pre-trained model, which outperforms models not trained on any data in those languages.
This gives valuable insights into the use of more languages of minimal multilingual data in English-centric pre-training.

\paragraph{LLaMA possesses better transfer ability across seen languages than BLOOM.}

The first subplot of Figure \ref{fig:transfer} shows the accuracy improvements from training in the three languages (EN, FR, ES) seen by all three models and testing in the same three languages (EN, FR, ES) on directly zero-shot testing in the three languages.
LLaMA demonstrates better or comparable multilingual transfer ability for all the training languages. Since LLaMA was trained on mostly English texts, it is natural to expect that it learns English data in finetuning better than multi-lingual models like BLOOM. This is consistent with the experimental result, where both minimum and maximum improvements for LLaMA are greater than those for BLOOM. 
Among the three models, Pythia has consistently lower improvements over zero-shot learning. We conjecture that the size of the English pre-training corpus has a positive correlation with a model's multilingual transfer ability. 
It is worth noting that the accuracy improvements achieved by all three models in the left one are higher than the second and third sub-figures. We argue that the common Latin script for the three languages might account for this significant increase. 
Although there are variations in collation, graphemes or phonetic values, the languages all involve the 26 most widespread letters with similar semantic information in their alphabets. This overlap of alphabets can contribute to the high effectiveness of fine-tuning in one language with the evaluation done in another.

\paragraph{Both English-centric models, i.e., LLaMA and Pythia, transfer better on rarely seen languages than BLOOM.}

As illustrated in the second subplot of Figure \ref{fig:transfer}, on the one hand, LLaMA exhibits more effective knowledge acquisition from Turkish (TR) and Greek (EL) data than BLOOM, which enhances its reasoning ability regardless of the language in which it is evaluated. This proves that pre-training with most data in one language can potentially benefit a model's ability to comprehend unfamiliar languages. 
On the other hand, Pythia emerges as the best-performing model when trained and evaluated on rarely seen languages by LLaMA and BLOOM. Considering the performance difference between Pythia and LLaMA, which are both English-centric models, we argue that the former's superiority can partially be attributed to the different language distributions of their pre-training dataset excluding English data. This suggests that even with fewer overall pre-training data, models can have a better transfer result after pre-training in the specific language. 



\paragraph{Language coverage in pre-training is still important for multilingual transfer.}
As illustrated in the third subplot of Figure \ref{fig:transfer}, we found that BLOOM overall performed the best, surpassing the other two models by a great margin. This is not surprising because BLOOM is trained on them while others are rarely. Pythia comes as the second, with LLaMA being the last. The superiority of Pythia over LLaMA can be attributed to the difference in their pre-train datasets. For Pythia, its dataset consists of 97.4\% of English data with the remaining for other languages, whereas for LLaMA, more than 99\% of its pre-train data is in English. Therefore, we suspect that a slightly more diverse pre-train dataset in languages benefits Pythia towards capturing linguistic universals. 

Finally, as illustrated in the fourth subplot of Figure \ref{fig:transfer}, we show that when training and evaluating in languages that LLaMA has seen but BLOOM hasn't, the test accuracies of LLaMA are significantly higher than the other two models, with Pythia being the second. 
This further suggests language coverage in pre-training is important for both multilingual models and English-centric models.

\subsection{Findings for RQ2}
\label{sec:rq2}
\textit{"RQ2: How does the source language used for downstream task finetuning affect multilingual reasoning on other languages?"}

\begin{figure}[t]
\centering
\includegraphics[width=5in]{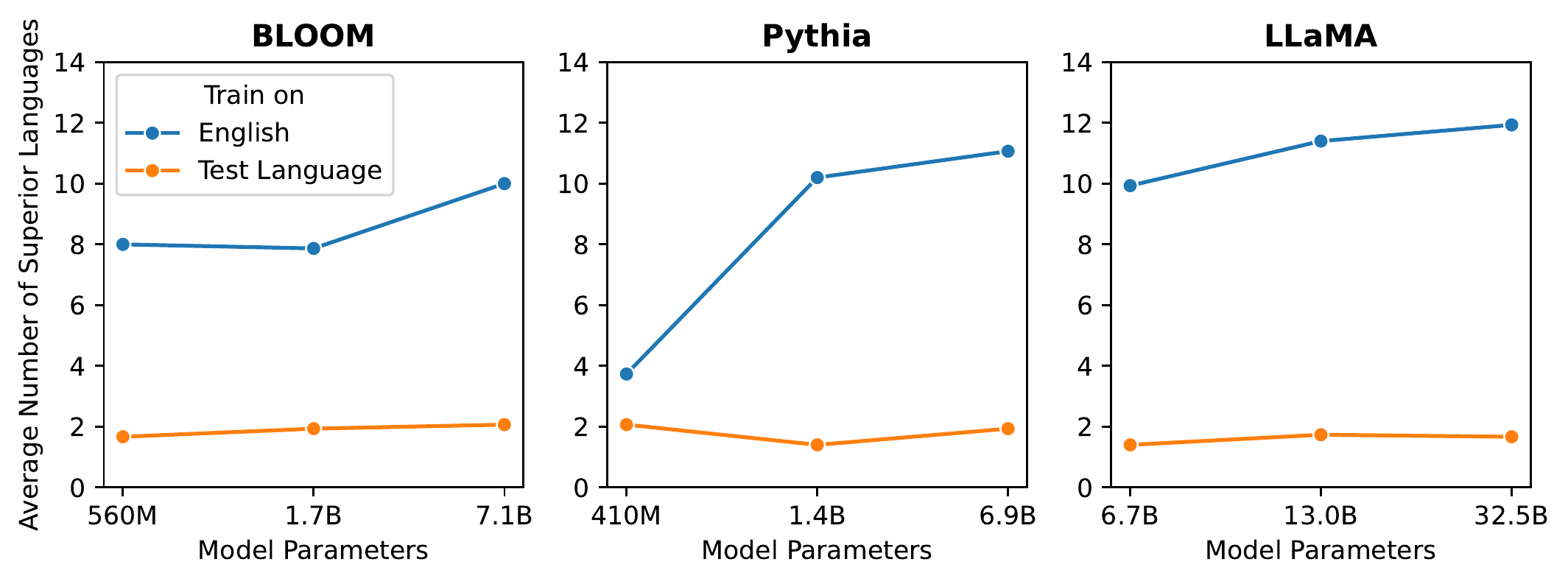}
\caption{Average number of superior training languages compared with English and the test language.
}
\label{fig:rq2.1}
\end{figure}

\paragraph{For both multilingual and English-centric models, English appears to be a less suitable source language when the model scales up.}
To investigate how the source language used for fine-tuning behaves on different models with different model sizes. We calculate the average number of superior source languages compared with English and the target language on the XNLI dataset. The values range from 0 to 14, indicating the certain source language (i.e., English or the target language) is from the best to the worst among the total 15 languages, respectively. We show the results in Figure~\ref{fig:rq2.1}.

As the model scales up, our experiments reveal that for all three models, there is a general increasing trend for the number of superior languages compared to English as model parameters grow. This observation can be attributed to the increasing capacity of the model, which enables it to capture more nuanced linguistic features. A possible explanation is as the increase of model capacity, the learning of other source languages becomes easier and consequently enhances the chances of identifying a more suitable source language other than English. These findings are applicable not only to multilingual models but also to both English-centric models.
Furthermore, we observe that under similar model parameters, the number of accuracy improvements over English fine-tuning for an English-centric model can be roughly equal (LLaMA-6.7B) or even higher (Pythia-6.9B) than a multilingual pre-trained model (BLOOM-7.1B). This can be counterintuitive because a model that knows English better is expected to learn better in English. 
However, our findings suggest otherwise, which highlights the need for further investigation in future research.

\paragraph{Training on target language may not be the best choice but can be a safe option.}

While training in the target language is not always the optimal choice, we find it consistently yields good performance. Based on Figure \ref{fig:rq2.1}, there are a small fraction of cases, with a number of approximately 2, where the accuracy difference is obtained by subtracting the accuracy of the model trained on each target language itself from trained on other languages, is positive. This finding suggests that incorporating target language data during training allows the model to better adapt to the specific characteristics of that language. 

\begin{figure}[t]
\centering
\includegraphics[width=5.3in]{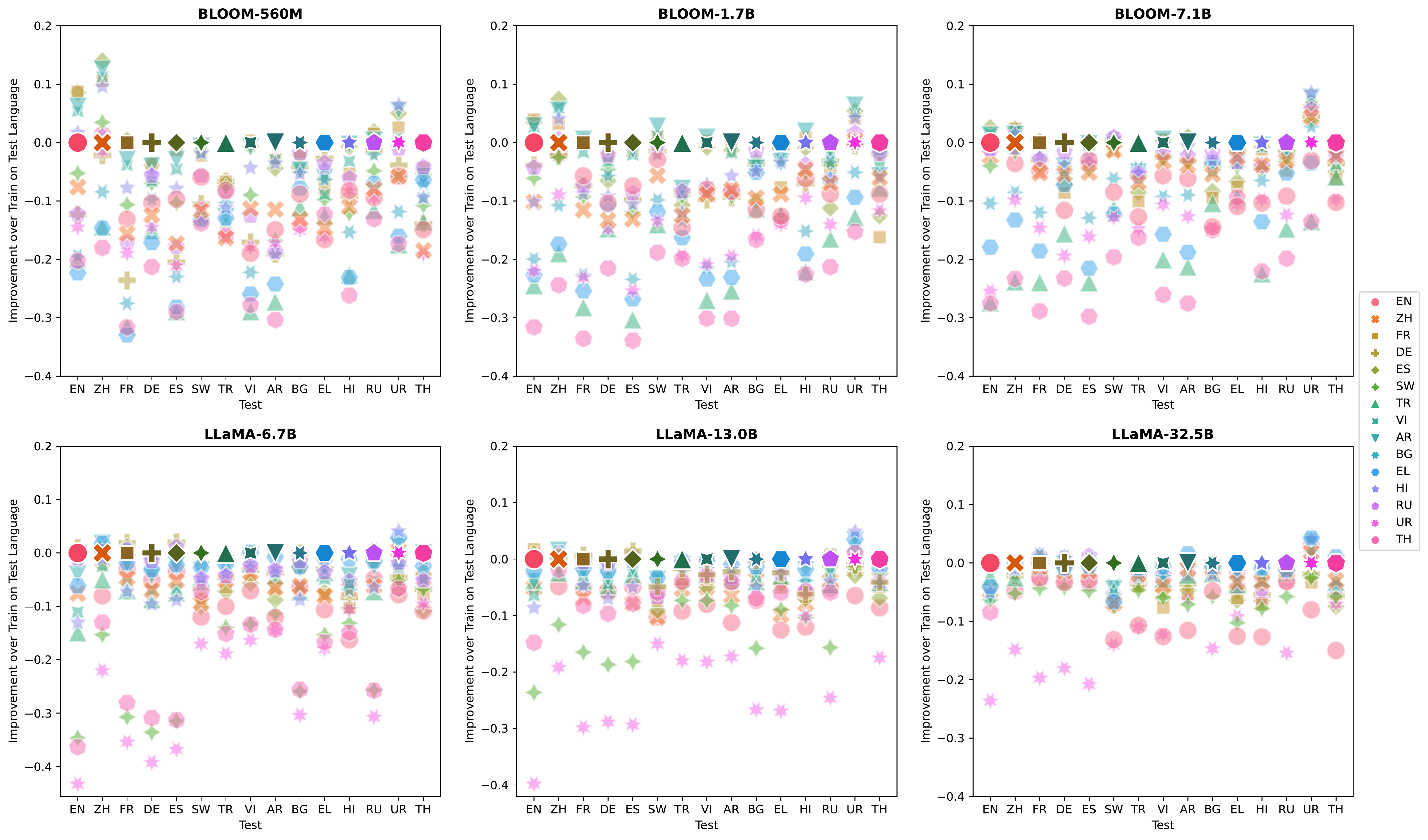}
\caption{Accuracy gain of BLOOMs and LLaMAs on test languages by subtracting the performance of models trained on each test language from those trained on other languages.
}
\label{fig:rq2.2}
\end{figure}

We further delve into each language to see if the on-average two superior languages are always the same for different models. To achieve this, we set the performance of the model trained on the target language as the baseline (0), and compute the relative performance gap of the model trained on each other source language. As shown in Figure~\ref{fig:rq2.2}, we find that such occurrences are primarily observed in Chinese (ZH), French (FR), Spanish (ES), and Urdu (UR), for LLaMA. While for BLOOM, they are mostly English (EN),  Chinese (ZH), and Urdu (UR). The results appear to be complex as they are not highly correlated with the language frequency observed during the pre-training stage. For instance, LLaMA has seen English, French, and Spanish, while BLOOM has seen English and Chinese. One possible explanation for this can be the distinctive language scripts used in Chinese (Chinese ideograms) and Urdu (Perso-Arabic), which may not be well-suited for acquiring knowledge related to reasoning.


\paragraph{Languages used in finetuning become increasingly irrelevant as an English-centric model scales up.}
In terms of Figure \ref{fig:rq2.2}, as the parameters of LLaMA grow, the distribution of Y-coordinates (i.e., accuracy improvements) becomes more concentrated around the line $y=0$, which corresponds to training and testing on the same language. 
Through a comparison between the LLaMA-6.7B and LLaMA-32.5B models, we find that the larger model not only exhibited fewer negative outliers, which were mostly associated with SW, UR, and TH as train languages, but also demonstrates significant accuracy improvements for other languages. As a result, the difference in accuracy between training on the target language and training on other languages is reduced when the model gets larger.
In contrast, we does not observe a clear trend with BLOOM as the model size increased from 560M to 7.1B. Additionally, we find the results on Pythia as shown in Appendix Figure~\ref{fig:on-english} to be less conclusive than those on LLaMA, and we attribute this to both the model size and the English-centric pre-training.

\subsection{Findings for RQ3}
\label{sec:rq3}
\textit{"RQ3: How does task type affect multilingual reasoning, e.g., will the reasoning ability be transferred better across languages in some reasoning tasks?"}

Previous work finds the transfer performance on `lower-level' tasks (e.g., POS-tagging, dependency parsing, and NER) to be better correlated with the syntactic similarity between languages, while `high-level' tasks (e.g., NLI and QA) rely more on other factors such as the size of pretraining corpora of the target language~\citep{lauscher-etal-2020-zero}. We are interested to see does transfer performance also differs in high-level reasoning tasks.

\begin{figure}[t]
\centering
\includegraphics[width=2.6in]{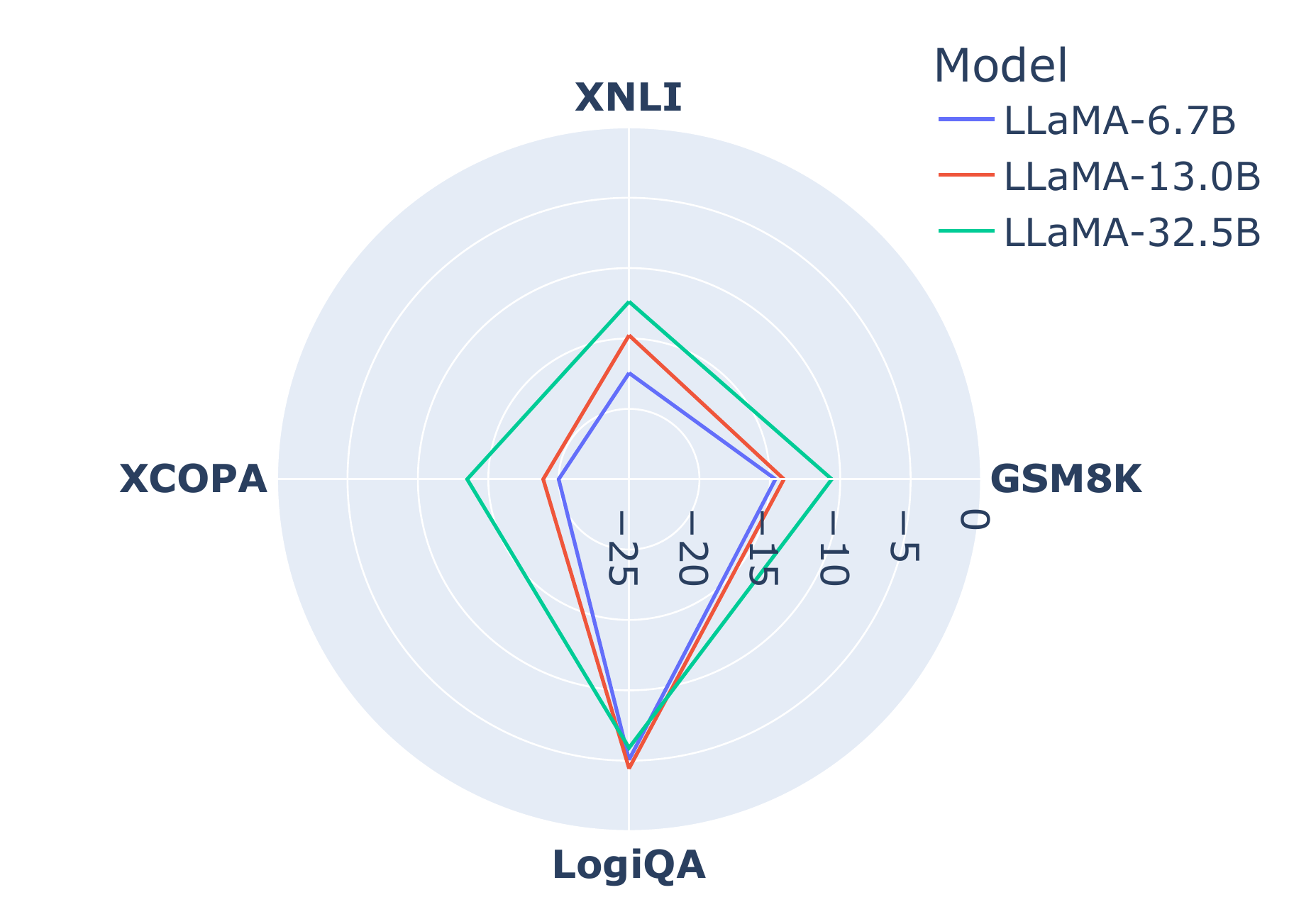}
\caption{Performance gap between the average accuracy on other languages (i.e., FR and ZH) and English using the Engligh-trained model. 0 refers to no performance gap, meaning the task ability transfers well from English to others.}
\label{fig:q3.1}
\end{figure}

\paragraph{Logical reasoning knowledge can be transferred better across languages than others, and such transferability on most tasks can be enhanced by scaling model size, even with a fixed English-centric pretraining corpus.}
To measure the multilingual reasoning transfer ability for different tasks, we calculate the performance gap between the average accuracy on other languages and English using the English-trained model. As different tasks may contain different test languages, to be consistent, we consider three languages, i.e., English, French, and Chinese for all the tasks in this experiment. A value of 0 refers to no performance gap, meaning the reasoning ability transfers well from English to others. We show the results on LLaMA with various model sizes in Figure~\ref{fig:q3.1}.
The results indicate that LogiQA, which focuses on logical reasoning, exhibits the highest transferability across all the model sizes considered. On the other hand, XNLI, which tests with natural language inference, and GSM8K, which tests arithmetic reasoning, demonstrate comparatively lower levels of effectiveness. Furthermore, the figure indicates that increasing the model size generally leads to improved performance across most of the tasks, suggesting that multilingual reasoning transferability can be enhanced by increasing the model size, even if the training corpus remains constant.
However, the results are fairly stable when the model scales up for LogiQA, with around 5\% lower than the performance of testing on English, suggesting that solely increasing the model size only improves the transfer ability to a certain amount. 

\begin{figure}[t]
\centering
\includegraphics[width=5.5in]{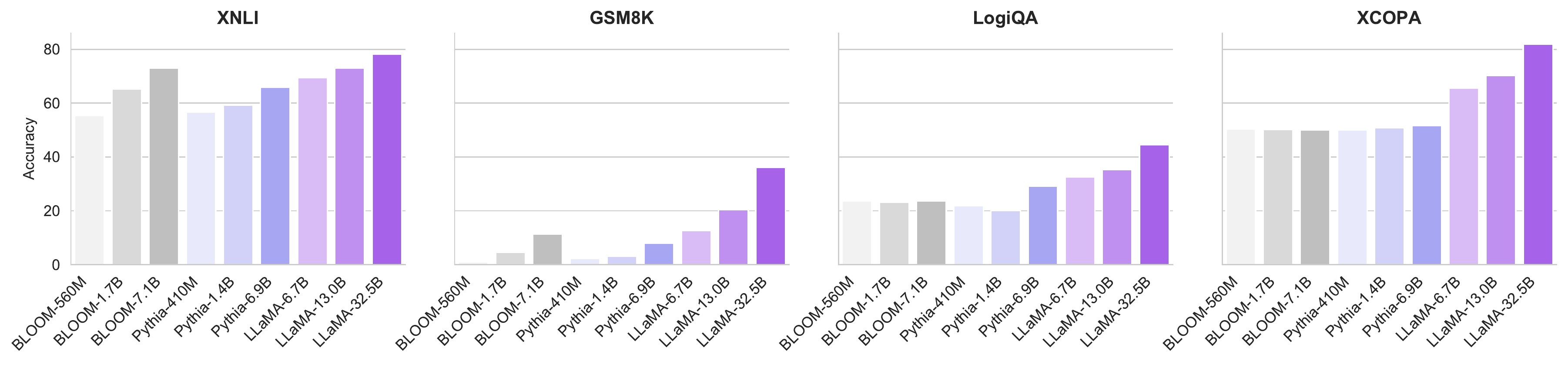}
\caption{Average accuracy on other languages (i.e., FR and ZH) of each model trained on English task data across the four tasks.}
\label{fig:q3.2}
\end{figure}

\paragraph{Multilingual pre-trained models fail on some multilingual reasoning tasks that English-centric models can handle.}
We further study the multilingual reasoning transfer ability of different types of models on the four tasks.
We show the average accuracy of the English-trained model when testing other languages in Figure~\ref{fig:q3.2}. Notably, BLOOM-7.1B failed on the LogiQA dataset, exhibiting a level of performance that was no better than random guessing, while both Pythia-6.9B and LLaMA-6.7B, two English-centric models, achieves better performance. This suggests that a multilingual model may not possess sufficient capability to learn certain types of reasoning tasks as an English-centric model does. Additionally, both BLOOM-7.1B and Pythia-6.9B failed on the XCOPA dataset. In contrast, LLaMA-7B performed significantly better on both of these tasks, highlighting the importance of considering the fundamental capabilities of a language model in the context of multilingual reasoning tasks.

\section{Conclusion}
In this work, we investigate the multilingual transfer capabilities of both multilingual pre-trained and English-centric models, on four multilingual reasoning tasks. Our findings suggest that English-centric models possess significant multilingual transferability. We also found that English may not be the most effective source language for English-centric models, and different types of reasoning tasks exhibit varying multilingual transfer abilities. 
These findings offer practical insights for both pre-training and fine-tuning of the multilingual and English-centric models. For example, including as many other languages as possible during pre-training, even with minimal amounts of data for each language, can be a cost-effective way to significantly improve the multilingual transfer capabilities of an English-centric model. In addition, the better adaptation ability to new languages on LLaMA also suggests injecting multilingual ability in the finetuning stage instead of the pre-training stage is also a viable way toward multilingual models. 
We hope that our study will inspire further investigations and advancements in the development of more effective multilingual models.

\section*{Limitation}
In this section, we discuss some potential limitations in our work. 
BLOOM and LLaMA, taken as representatives for language versatilist and specialist respectively, might not be strictly comparable because they were trained on different quantities of data. Hence, the results derived in our paper could tend to favor LLaMA which was pre-trained on more data considering all languages. To alleviate this inequality, we have conducted experiments on Pythia with a smaller pre-train dataset. If the corresponding result is still better than that of BLOOM, then we can conclude with stronger confidence that the specialist approach is superior. Nevertheless, noting that the quality of pre-train datasets can also vary, which makes Pythia and BLOOM still not strictly comparable. We acknowledge such possible deviations in the amount and quality of the pre-training corpus for the three models, and we recommend that future research pays more attention to it.
In addition, we only evaluated the performance of supervised task fine-tuning in our study. In future work, it would be worthwhile to consider other learning paradigms such as in-context learning~\citep{brown2020language}.

\medskip
{
\small
\bibliography{ref}

\begin{thebibliography}{45}
\providecommand{\natexlab}[1]{#1}
\providecommand{\url}[1]{\texttt{#1}}
\expandafter\ifx\csname urlstyle\endcsname\relax
  \providecommand{\doi}[1]{doi: #1}\else
  \providecommand{\doi}{doi: \begingroup \urlstyle{rm}\Url}\fi

\bibitem[Artetxe et~al.(2020)Artetxe, Ruder, and Yogatama]{artetxe2020}
Mikel Artetxe, Sebastian Ruder, and Dani Yogatama.
\newblock On the cross-lingual transferability of monolingual representations.
\newblock In \emph{Proceedings of the 58th Annual Meeting of the Association
  for Computational Linguistics}, pages 4623--4637, Online, 2020. Association
  for Computational Linguistics.
\newblock \doi{10.18653/v1/2020.acl-main.421}.
\newblock URL \url{https://aclanthology.org/2020.acl-main.421}.

\bibitem[Biderman et~al.(2023)Biderman, Schoelkopf, Anthony, Bradley, O'Brien,
  Hallahan, Khan, Purohit, Prashanth, Raff, Skowron, Sutawika, and van~der
  Wal]{biderman2023pythia}
Stella Biderman, Hailey Schoelkopf, Quentin Anthony, Herbie Bradley, Kyle
  O'Brien, Eric Hallahan, Mohammad~Aflah Khan, Shivanshu Purohit, USVSN~Sai
  Prashanth, Edward Raff, Aviya Skowron, Lintang Sutawika, and Oskar van~der
  Wal.
\newblock Pythia: A suite for analyzing large language models across training
  and scaling, 2023.

\bibitem[Black et~al.(2021)Black, Gao, Wang, Leahy, and Biderman]{gpt-neo}
Sid Black, Leo Gao, Phil Wang, Connor Leahy, and Stella Biderman.
\newblock {GPT-Neo: Large Scale Autoregressive Language Modeling with
  Mesh-Tensorflow}, March 2021.
\newblock URL \url{https://doi.org/10.5281/zenodo.5297715}.
\newblock {If you use this software, please cite it using these metadata.}

\bibitem[Black et~al.(2022)Black, Biderman, Hallahan, Anthony, Gao, Golding,
  He, Leahy, McDonell, Phang, Pieler, Prashanth, Purohit, Reynolds, Tow, Wang,
  and Weinbach]{gpt-neox-20b}
Sid Black, Stella Biderman, Eric Hallahan, Quentin Anthony, Leo Gao, Laurence
  Golding, Horace He, Connor Leahy, Kyle McDonell, Jason Phang, Michael Pieler,
  USVSN~Sai Prashanth, Shivanshu Purohit, Laria Reynolds, Jonathan Tow, Ben
  Wang, and Samuel Weinbach.
\newblock {GPT-NeoX-20B}: An open-source autoregressive language model.
\newblock In \emph{Proceedings of the ACL Workshop on Challenges \&
  Perspectives in Creating Large Language Models}, 2022.
\newblock URL \url{https://arxiv.org/abs/2204.06745}.

\bibitem[Brown et~al.(2020)Brown, Mann, Ryder, Subbiah, Kaplan, Dhariwal,
  Neelakantan, Shyam, Sastry, Askell, et~al.]{brown2020language}
Tom Brown, Benjamin Mann, Nick Ryder, Melanie Subbiah, Jared~D Kaplan, Prafulla
  Dhariwal, Arvind Neelakantan, Pranav Shyam, Girish Sastry, Amanda Askell,
  et~al.
\newblock Language models are few-shot learners.
\newblock \emph{Advances in neural information processing systems},
  33:\penalty0 1877--1901, 2020.

\bibitem[Chomsky(1981)]{chomsky1981naturalistic}
Noam Chomsky.
\newblock A naturalistic approach to language and cognition.
\newblock \emph{Cognition and Brain Theory}, 4\penalty0 (1):\penalty0 3--22,
  1981.

\bibitem[Chowdhery et~al.(2022)Chowdhery, Narang, Devlin, Bosma, Mishra,
  Roberts, Barham, Chung, Sutton, Gehrmann, et~al.]{chowdhery2022palm}
Aakanksha Chowdhery, Sharan Narang, Jacob Devlin, Maarten Bosma, Gaurav Mishra,
  Adam Roberts, Paul Barham, Hyung~Won Chung, Charles Sutton, Sebastian
  Gehrmann, et~al.
\newblock Palm: Scaling language modeling with pathways.
\newblock \emph{arXiv preprint arXiv:2204.02311}, 2022.

\bibitem[Cobbe et~al.(2021)Cobbe, Kosaraju, Bavarian, Chen, Jun, Kaiser,
  Plappert, Tworek, Hilton, Nakano, et~al.]{cobbe2021training}
Karl Cobbe, Vineet Kosaraju, Mohammad Bavarian, Mark Chen, Heewoo Jun, Lukasz
  Kaiser, Matthias Plappert, Jerry Tworek, Jacob Hilton, Reiichiro Nakano,
  et~al.
\newblock Training verifiers to solve math word problems.
\newblock \emph{arXiv preprint arXiv:2110.14168}, 2021.

\bibitem[Conneau et~al.(2018{\natexlab{a}})Conneau, Rinott, Lample, Williams,
  Bowman, Schwenk, and Stoyanov]{conneau2018}
Alexis Conneau, Ruty Rinott, Guillaume Lample, Adina Williams, Samuel Bowman,
  Holger Schwenk, and Veselin Stoyanov.
\newblock {XNLI}: Evaluating cross-lingual sentence representations.
\newblock In \emph{Proceedings of the 2018 Conference on Empirical Methods in
  Natural Language Processing}, pages 2475--2485, Brussels, Belgium,
  2018{\natexlab{a}}. Association for Computational Linguistics.
\newblock \doi{10.18653/v1/D18-1269}.
\newblock URL \url{https://aclanthology.org/D18-1269}.

\bibitem[Conneau et~al.(2018{\natexlab{b}})Conneau, Rinott, Lample, Williams,
  Bowman, Schwenk, and Stoyanov]{conneau2018xnli}
Alexis Conneau, Ruty Rinott, Guillaume Lample, Adina Williams, Samuel Bowman,
  Holger Schwenk, and Veselin Stoyanov.
\newblock Xnli: Evaluating cross-lingual sentence representations.
\newblock In \emph{Proceedings of the 2018 Conference on Empirical Methods in
  Natural Language Processing}, pages 2475--2485, 2018{\natexlab{b}}.

\bibitem[Conneau et~al.(2020)Conneau, Khandelwal, Goyal, Chaudhary, Wenzek,
  Guzm{\'a}n, Grave, Ott, Zettlemoyer, and Stoyanov]{conneau2020}
Alexis Conneau, Kartikay Khandelwal, Naman Goyal, Vishrav Chaudhary, Guillaume
  Wenzek, Francisco Guzm{\'a}n, Edouard Grave, Myle Ott, Luke Zettlemoyer, and
  Veselin Stoyanov.
\newblock Unsupervised cross-lingual representation learning at scale.
\newblock In \emph{Proceedings of the 58th Annual Meeting of the Association
  for Computational Linguistics}, pages 8440--8451, Online, 2020. Association
  for Computational Linguistics.
\newblock \doi{10.18653/v1/2020.acl-main.747}.
\newblock URL \url{https://aclanthology.org/2020.acl-main.747}.

\bibitem[Dettmers et~al.(2022)Dettmers, Lewis, Belkada, and
  Zettlemoyer]{dettmers2022llmint8}
Tim Dettmers, Mike Lewis, Younes Belkada, and Luke Zettlemoyer.
\newblock Llm.int8(): 8-bit matrix multiplication for transformers at scale.
\newblock \emph{arXiv preprint arXiv:2208.07339}, 2022.

\bibitem[Devlin et~al.(2019)Devlin, Chang, Lee, and Toutanova]{devlin2019bert}
Jacob Devlin, Ming-Wei Chang, Kenton Lee, and Kristina Toutanova.
\newblock Bert: Pre-training of deep bidirectional transformers for language
  understanding.
\newblock In \emph{Proceedings of the 2019 Conference of the North American
  Chapter of the Association for Computational Linguistics: Human Language
  Technologies, Volume 1 (Long and Short Papers)}, pages 4171--4186, 2019.

\bibitem[Gao et~al.(2020)Gao, Biderman, Black, Golding, Hoppe, Foster, Phang,
  He, Thite, Nabeshima, et~al.]{gao2020pile}
Leo Gao, Stella Biderman, Sid Black, Laurence Golding, Travis Hoppe, Charles
  Foster, Jason Phang, Horace He, Anish Thite, Noa Nabeshima, et~al.
\newblock The pile: An 800gb dataset of diverse text for language modeling.
\newblock \emph{arXiv preprint arXiv:2101.00027}, 2020.

\bibitem[Hernandez et~al.(2021)Hernandez, Kaplan, Henighan, and
  McCandlish]{hernandez2021scaling}
Danny Hernandez, Jared Kaplan, Tom Henighan, and Sam McCandlish.
\newblock Scaling laws for transfer.
\newblock \emph{arXiv preprint arXiv:2102.01293}, 2021.

\bibitem[Hu et~al.(2022)Hu, Wallis, Allen-Zhu, Li, Wang, Wang, Chen,
  et~al.]{hulora}
Edward~J Hu, Phillip Wallis, Zeyuan Allen-Zhu, Yuanzhi Li, Shean Wang, Lu~Wang,
  Weizhu Chen, et~al.
\newblock Lora: Low-rank adaptation of large language models.
\newblock In \emph{International Conference on Learning Representations}, 2022.

\bibitem[Hu et~al.(2020)Hu, Ruder, Siddhant, Neubig, Firat, and
  Johnson]{hu2020xtreme}
Junjie Hu, Sebastian Ruder, Aditya Siddhant, Graham Neubig, Orhan Firat, and
  Melvin Johnson.
\newblock Xtreme: A massively multilingual multi-task benchmark for evaluating
  cross-lingual generalisation.
\newblock In \emph{International Conference on Machine Learning}, pages
  4411--4421. PMLR, 2020.

\bibitem[Lauren{\c{c}}on et~al.(2022)Lauren{\c{c}}on, Saulnier, Wang, Akiki,
  Villanova~del Moral, Le~Scao, Von~Werra, Mou, Gonz{\'a}lez~Ponferrada,
  Nguyen, et~al.]{laurenccon2022bigscience}
Hugo Lauren{\c{c}}on, Lucile Saulnier, Thomas Wang, Christopher Akiki, Albert
  Villanova~del Moral, Teven Le~Scao, Leandro Von~Werra, Chenghao Mou, Eduardo
  Gonz{\'a}lez~Ponferrada, Huu Nguyen, et~al.
\newblock The bigscience roots corpus: A 1.6 tb composite multilingual dataset.
\newblock \emph{Advances in Neural Information Processing Systems},
  35:\penalty0 31809--31826, 2022.

\bibitem[Lauscher et~al.(2020)Lauscher, Ravishankar, Vuli{\'c}, and
  Glava{\v{s}}]{lauscher-etal-2020-zero}
Anne Lauscher, Vinit Ravishankar, Ivan Vuli{\'c}, and Goran Glava{\v{s}}.
\newblock From zero to hero: {O}n the limitations of zero-shot language
  transfer with multilingual {T}ransformers.
\newblock In \emph{Proceedings of the 2020 Conference on Empirical Methods in
  Natural Language Processing (EMNLP)}, pages 4483--4499, Online, November
  2020. Association for Computational Linguistics.
\newblock \doi{10.18653/v1/2020.emnlp-main.363}.
\newblock URL \url{https://aclanthology.org/2020.emnlp-main.363}.

\bibitem[Legate and Yang(2002)]{legate2002empirical}
Julie~Anne Legate and Charles~D Yang.
\newblock Empirical re-assessment of stimulus poverty arguments.
\newblock \emph{The Linguistic Review}, 19\penalty0 (1-2):\penalty0 151--162,
  2002.

\bibitem[Lin et~al.(2021)Lin, Mihaylov, Artetxe, Wang, Chen, Simig, Ott, Goyal,
  Bhosale, Du, Pasunuru, Shleifer, Koura, Chaudhary, O'Horo, Wang, Zettlemoyer,
  Kozareva, Diab, Stoyanov, and Li]{lin2022e}
Xi~Victoria Lin, Todor Mihaylov, Mikel Artetxe, Tianlu Wang, Shuohui Chen,
  Daniel Simig, Myle Ott, Naman Goyal, Shruti Bhosale, Jingfei Du, Ramakanth
  Pasunuru, Sam Shleifer, Punit~Singh Koura, Vishrav Chaudhary, Brian O'Horo,
  Jeff Wang, Luke Zettlemoyer, Zornitsa Kozareva, Mona Diab, Veselin Stoyanov,
  and Xian Li.
\newblock Few-shot {{Learning}} with {{Multilingual Language Models}}, 2021.
\newblock URL \url{https://arxiv.org/abs/2112.10668}.

\bibitem[Liu et~al.(2021)Liu, Cui, Liu, Huang, Wang, and Zhang]{liu2021logiqa}
Jian Liu, Leyang Cui, Hanmeng Liu, Dandan Huang, Yile Wang, and Yue Zhang.
\newblock Logiqa: a challenge dataset for machine reading comprehension with
  logical reasoning.
\newblock In \emph{Proceedings of the Twenty-Ninth International Conference on
  International Joint Conferences on Artificial Intelligence}, pages
  3622--3628, 2021.

\bibitem[Lu et~al.(2021)Lu, Grover, Abbeel, and Mordatch]{lu2021pretrained}
Kevin Lu, Aditya Grover, Pieter Abbeel, and Igor Mordatch.
\newblock Pretrained transformers as universal computation engines.
\newblock \emph{arXiv preprint arXiv:2103.05247}, 1, 2021.

\bibitem[Muennighoff et~al.(2022)Muennighoff, Wang, Sutawika, Roberts,
  Biderman, Scao, Bari, Shen, Yong, Schoelkopf, Tang, Radev, Aji, Almubarak,
  Albanie, Alyafeai, Webson, Raff, and Raffel]{muennighoff2022}
Niklas Muennighoff, Thomas Wang, Lintang Sutawika, Adam Roberts, Stella
  Biderman, Teven~Le Scao, M.~Saiful Bari, Sheng Shen, Zheng-Xin Yong, Hailey
  Schoelkopf, Xiangru Tang, Dragomir Radev, Alham~Fikri Aji, Khalid Almubarak,
  Samuel Albanie, Zaid Alyafeai, Albert Webson, Edward Raff, and Colin Raffel.
\newblock Crosslingual {{Generalization}} through {{Multitask Finetuning}},
  2022.
\newblock URL \url{https://arxiv.org/abs/2211.01786}.

\bibitem[Papadimitriou and Jurafsky(2020)]{papadimitriou2020pretraining}
Isabel Papadimitriou and Dan Jurafsky.
\newblock Pretraining on non-linguistic structure as a tool for analyzing
  learning bias in language models.
\newblock \emph{arXiv preprint arXiv:2004.14601}, 2020.

\bibitem[Pfeiffer et~al.(2022)Pfeiffer, Goyal, Lin, Li, Cross, Riedel, and
  Artetxe]{pfeiffer2022}
Jonas Pfeiffer, Naman Goyal, Xi~Lin, Xian Li, James Cross, Sebastian Riedel,
  and Mikel Artetxe.
\newblock Lifting the curse of multilinguality by pre-training modular
  transformers.
\newblock In \emph{Proceedings of the 2022 Conference of the North American
  Chapter of the Association for Computational Linguistics: Human Language
  Technologies}, pages 3479--3495, Seattle, United States, 2022. Association
  for Computational Linguistics.
\newblock \doi{10.18653/v1/2022.naacl-main.255}.
\newblock URL \url{https://aclanthology.org/2022.naacl-main.255}.

\bibitem[Pires et~al.(2019)Pires, Schlinger, and Garrette]{pires2019}
Telmo Pires, Eva Schlinger, and Dan Garrette.
\newblock How multilingual is multilingual {BERT}?
\newblock In \emph{Proceedings of the 57th Annual Meeting of the Association
  for Computational Linguistics}, pages 4996--5001, Florence, Italy, 2019.
  Association for Computational Linguistics.
\newblock \doi{10.18653/v1/P19-1493}.
\newblock URL \url{https://aclanthology.org/P19-1493}.

\bibitem[Ponti et~al.(2020)Ponti, Glava{\v{s}}, Majewska, Liu, Vuli{\'c}, and
  Korhonen]{ponti2020xcopa}
Edoardo~Maria Ponti, Goran Glava{\v{s}}, Olga Majewska, Qianchu Liu, Ivan
  Vuli{\'c}, and Anna Korhonen.
\newblock Xcopa: A multilingual dataset for causal commonsense reasoning.
\newblock In \emph{Proceedings of the 2020 Conference on Empirical Methods in
  Natural Language Processing (EMNLP)}, pages 2362--2376, 2020.

\bibitem[Ri and Tsuruoka(2022)]{ri2022pretraining}
Ryokan Ri and Yoshimasa Tsuruoka.
\newblock Pretraining with artificial language: Studying transferable knowledge
  in language models.
\newblock In \emph{Proceedings of the 60th Annual Meeting of the Association
  for Computational Linguistics (Volume 1: Long Papers)}, pages 7302--7315,
  2022.

\bibitem[Roemmele et~al.(2011)Roemmele, Bejan, and Gordon]{roemmele2011choice}
Melissa Roemmele, Cosmin~Adrian Bejan, and Andrew~S Gordon.
\newblock Choice of plausible alternatives: An evaluation of commonsense causal
  reasoning.
\newblock In \emph{AAAI spring symposium: logical formalizations of commonsense
  reasoning}, pages 90--95, 2011.

\bibitem[Sanh et~al.(2021)Sanh, Webson, Raffel, Bach, Sutawika, Alyafeai,
  Chaffin, Stiegler, Raja, Dey, et~al.]{sanhmultitask}
Victor Sanh, Albert Webson, Colin Raffel, Stephen Bach, Lintang Sutawika, Zaid
  Alyafeai, Antoine Chaffin, Arnaud Stiegler, Arun Raja, Manan Dey, et~al.
\newblock Multitask prompted training enables zero-shot task generalization.
\newblock In \emph{International Conference on Learning Representations}, 2021.

\bibitem[Scao et~al.(2022)Scao, Fan, Akiki, Pavlick, Ili{\'c}, Hesslow,
  Castagn{\'e}, Luccioni, Yvon, Gall{\'e}, et~al.]{scao2022bloom}
Teven~Le Scao, Angela Fan, Christopher Akiki, Ellie Pavlick, Suzana Ili{\'c},
  Daniel Hesslow, Roman Castagn{\'e}, Alexandra~Sasha Luccioni, Fran{\c{c}}ois
  Yvon, Matthias Gall{\'e}, et~al.
\newblock Bloom: A 176b-parameter open-access multilingual language model.
\newblock \emph{arXiv preprint arXiv:2211.05100}, 2022.

\bibitem[Shi et~al.(2022)Shi, Suzgun, Freitag, Wang, Srivats, Vosoughi, Chung,
  Tay, Ruder, Zhou, Das, and Wei]{shi2022b}
Freda Shi, Mirac Suzgun, Markus Freitag, Xuezhi Wang, Suraj Srivats, Soroush
  Vosoughi, Hyung~Won Chung, Yi~Tay, Sebastian Ruder, Denny Zhou, Dipanjan Das,
  and Jason Wei.
\newblock Language {{Models}} are {{Multilingual Chain-of-Thought Reasoners}},
  2022.
\newblock URL \url{https://arxiv.org/abs/2210.03057}.

\bibitem[Shliazhko et~al.(2022)Shliazhko, Fenogenova, Tikhonova, Mikhailov,
  Kozlova, and Shavrina]{shliazhko2022}
Oleh Shliazhko, Alena Fenogenova, Maria Tikhonova, Vladislav Mikhailov,
  Anastasia Kozlova, and Tatiana Shavrina.
\newblock {{mGPT}}: {{Few-Shot Learners Go Multilingual}}, 2022.
\newblock URL \url{https://arxiv.org/abs/2204.07580}.

\bibitem[Touvron et~al.(2023)Touvron, Lavril, Izacard, Martinet, Lachaux,
  Lacroix, Rozi{\`e}re, Goyal, Hambro, Azhar, et~al.]{touvron2023llama}
Hugo Touvron, Thibaut Lavril, Gautier Izacard, Xavier Martinet, Marie-Anne
  Lachaux, Timoth{\'e}e Lacroix, Baptiste Rozi{\`e}re, Naman Goyal, Eric
  Hambro, Faisal Azhar, et~al.
\newblock Llama: Open and efficient foundation language models.
\newblock \emph{arXiv preprint arXiv:2302.13971}, 2023.

\bibitem[Wang and Komatsuzaki(2021)]{gpt-j}
Ben Wang and Aran Komatsuzaki.
\newblock {GPT-J-6B: A 6 Billion Parameter Autoregressive Language Model}.
\newblock \url{https://github.com/kingoflolz/mesh-transformer-jax}, May 2021.

\bibitem[Wang et~al.(2020)Wang, Lipton, and Tsvetkov]{wang-etal-2020-negative}
Zirui Wang, Zachary~C. Lipton, and Yulia Tsvetkov.
\newblock On negative interference in multilingual models: Findings and a
  meta-learning treatment.
\newblock In \emph{Proceedings of the 2020 Conference on Empirical Methods in
  Natural Language Processing (EMNLP)}, pages 4438--4450, Online, November
  2020. Association for Computational Linguistics.
\newblock \doi{10.18653/v1/2020.emnlp-main.359}.
\newblock URL \url{https://aclanthology.org/2020.emnlp-main.359}.

\bibitem[Wei et~al.(2021)Wei, Bosma, Zhao, Guu, Yu, Lester, Du, Dai, and
  Le]{weifinetuned}
Jason Wei, Maarten Bosma, Vincent Zhao, Kelvin Guu, Adams~Wei Yu, Brian Lester,
  Nan Du, Andrew~M Dai, and Quoc~V Le.
\newblock Finetuned language models are zero-shot learners.
\newblock In \emph{International Conference on Learning Representations}, 2021.

\bibitem[Williams et~al.(2018)Williams, Nangia, and Bowman]{williams2018broad}
Adina Williams, Nikita Nangia, and Samuel Bowman.
\newblock A broad-coverage challenge corpus for sentence understanding through
  inference.
\newblock In \emph{Proceedings of the 2018 Conference of the North American
  Chapter of the Association for Computational Linguistics: Human Language
  Technologies, Volume 1 (Long Papers)}, pages 1112--1122, 2018.

\bibitem[Wu and Dredze(2019)]{wu-dredze-2019-beto}
Shijie Wu and Mark Dredze.
\newblock Beto, bentz, becas: The surprising cross-lingual effectiveness of
  {BERT}.
\newblock In \emph{Proceedings of the 2019 Conference on Empirical Methods in
  Natural Language Processing and the 9th International Joint Conference on
  Natural Language Processing (EMNLP-IJCNLP)}, pages 833--844, Hong Kong,
  China, November 2019. Association for Computational Linguistics.
\newblock \doi{10.18653/v1/D19-1077}.
\newblock URL \url{https://aclanthology.org/D19-1077}.

\bibitem[Wu et~al.(2023)Wu, Wang, Ye, Feng, Xu, Qiao, and Wu]{wu2023openicl}
Zhenyu Wu, YaoXiang Wang, Jiacheng Ye, Jiangtao Feng, Jingjing Xu, Yu~Qiao, and
  Zhiyong Wu.
\newblock Openicl: An open-source framework for in-context learning.
\newblock \emph{arXiv preprint arXiv:2303.02913}, 2023.

\bibitem[Xue et~al.(2021)Xue, Constant, Roberts, Kale, Al-Rfou, Siddhant,
  Barua, and Raffel]{xue2021}
Linting Xue, Noah Constant, Adam Roberts, Mihir Kale, Rami Al-Rfou, Aditya
  Siddhant, Aditya Barua, and Colin Raffel.
\newblock m{T}5: A massively multilingual pre-trained text-to-text transformer.
\newblock In \emph{Proceedings of the 2021 Conference of the North American
  Chapter of the Association for Computational Linguistics: Human Language
  Technologies}, pages 483--498, Online, 2021. Association for Computational
  Linguistics.
\newblock \doi{10.18653/v1/2021.naacl-main.41}.
\newblock URL \url{https://aclanthology.org/2021.naacl-main.41}.

\bibitem[Ye et~al.(2022)Ye, Gao, Li, Xu, Feng, Wu, Yu, and Kong]{ye2022zerogen}
Jiacheng Ye, Jiahui Gao, Qintong Li, Hang Xu, Jiangtao Feng, Zhiyong Wu, Tao
  Yu, and Lingpeng Kong.
\newblock Zerogen: Efficient zero-shot learning via dataset generation.
\newblock \emph{arXiv preprint arXiv:2202.07922}, 2022.

\bibitem[Zhang et~al.(2022)Zhang, Roller, Goyal, Artetxe, Chen, Chen, Dewan,
  Diab, Li, Lin, et~al.]{zhang2022opt}
Susan Zhang, Stephen Roller, Naman Goyal, Mikel Artetxe, Moya Chen, Shuohui
  Chen, Christopher Dewan, Mona Diab, Xian Li, Xi~Victoria Lin, et~al.
\newblock Opt: Open pre-trained transformer language models.
\newblock \emph{arXiv preprint arXiv:2205.01068}, 2022.

\bibitem[Zhu et~al.(2023)Zhu, Liu, Dong, Xu, Kong, Chen, Li, and
  Huang]{zhu2023multilingual}
Wenhao Zhu, Hongyi Liu, Qingxiu Dong, Jingjing Xu, Lingpeng Kong, Jiajun Chen,
  Lei Li, and Shujian Huang.
\newblock Multilingual machine translation with large language models:
  Empirical results and analysis.
\newblock \emph{arXiv preprint arXiv:2304.04675}, 2023.

\end{thebibliography}
}

\appendix

\section{Datasets and Templates}
\begin{table}[h]
\caption{Number of training and test instances for each dataset, as well as the templates used during fine-tuning and inference.}
\label{tab:prompts}
\scalebox{0.8}{
\begin{tabular}{lll}
\toprule
\textbf{} & \textbf{\#Train/\#Test} & \textbf{Template} \\
\midrule
\textbf{XNLI} & 9,000/5010 & \begin{tabular}[c]{@{}l@{}}Question:\\ \{premise\} Based on the previous passage, is it true that "\{hypothesis\}"? Yes, No, or Maybe?\\ \\ Answer:\\ \{output\}\end{tabular} \\
\midrule
\textbf{XCOPA} & 400/500 & \begin{tabular}[c]{@{}l@{}}Question:\\ \{premise\} Based on the previous passage, choose the most reasonable \{cause | effect\}.\\ A:\{choice1\}\\ B:\{choice2\}\\ \\ Answer:\\ \{output\}\end{tabular} \\
\midrule
\textbf{LogiQA} & 7,376/651 & \begin{tabular}[c]{@{}l@{}}Question:\\ \{context\} \{question\}\\ A: \{choice1\}\\ B: \{choice2\}\\ C: \{choice3\}\\ D: \{choice4\}\\ \\ Answer:\\ \{output\}\end{tabular} \\
\midrule
\textbf{GSM8K} & 7,473/250 & \begin{tabular}[c]{@{}l@{}}Question:\\ \{input\}\\ \\ Answer:\\ \{output\}\end{tabular} \\
\bottomrule
\end{tabular}}
\end{table}
We show the number of instances and the template used in each dataset in Table~\ref{tab:prompts}.

\section{Detailed Results}

\setlength{\tabcolsep}{1mm}
\begin{table}[ht]
\caption{Detailed results of BLOOM on XNLI dataset.}
\centering
\label{tab:bloom-xnli}
\scalebox{0.75}{
\begin{tabular}{llccccccccccccccccc}
\toprule
\multirow{2}{*}{\textbf{Model}} & \multicolumn{1}{c}{\multirow{2}{*}{\textbf{Train}}} & \multicolumn{15}{c}{\textbf{Test}} & \multicolumn{1}{c}{\multirow{2}{*}{\textbf{Average}}} \\
 & \multicolumn{1}{c}{} & \textbf{ar} & \textbf{bg} & \textbf{de} & \textbf{el} & \textbf{en} & \textbf{es} & \textbf{fr} & \textbf{hi} & \textbf{ru} & \textbf{sw} & \textbf{th} & \textbf{tr} & \textbf{ur} & \textbf{vi} & \textbf{zh} & \multicolumn{1}{c}{} \\
 \midrule
\multirow{17}{*}{\textbf{BLOOM-560M}} & \textbf{Zero-shot} & 33.29 & 34.53 & 34.81 & 34.61 & 35.01 & 34.53 & 33.99 & 33.71 & 34.71 & 33.79 & 34.05 & 33.49 & 33.17 & 34.03 & 33.73 & 34.10 \\
 & \textbf{Average} & 50.80 & 50.21 & 50.92 & 48.22 & 55.37 & 56.03 & 54.58 & 52.01 & 52.00 & 45.19 & 44.55 & 41.44 & 47.36 & 53.38 & 52.59 & 50.31 \\
 & \textbf{ar} & 64.97 & 52.08 & 56.01 & 50.74 & 68.42 & 67.92 & 67.43 & 61.80 & 56.53 & 51.54 & 48.24 & 39.00 & 57.21 & 65.83 & 64.83 & 58.17 \\
 & \textbf{bg} & 45.97 & 56.83 & 50.32 & 50.46 & 42.83 & 47.92 & 42.83 & 46.79 & 54.17 & 39.34 & 47.11 & 41.38 & 39.84 & 44.29 & 43.87 & 46.26 \\
 & \textbf{de} & 45.89 & 52.67 & 60.02 & 54.05 & 49.96 & 50.34 & 46.77 & 52.95 & 54.73 & 41.68 & 40.14 & 44.99 & 47.80 & 49.42 & 50.10 & 49.43 \\
 & \textbf{el} & 40.74 & 49.02 & 42.89 & 56.75 & 39.94 & 42.65 & 37.45 & 39.02 & 49.16 & 39.00 & 47.68 & 38.74 & 35.61 & 40.58 & 37.64 & 42.46 \\
 & \textbf{en} & 50.06 & 48.02 & 49.68 & 40.14 & 62.26 & 61.16 & 57.25 & 53.89 & 46.77 & 46.29 & 39.16 & 43.49 & 45.93 & 47.54 & 53.57 & 49.68 \\
 & \textbf{es} & 60.58 & 54.61 & 56.15 & 50.50 & 70.86 & 70.86 & 70.80 & 61.42 & 58.02 & 50.80 & 48.72 & 44.65 & 56.71 & 65.85 & 66.31 & 59.12 \\
 & \textbf{fr} & 62.10 & 55.19 & 56.47 & 48.82 & 71.04 & 69.58 & 70.36 & 56.75 & 57.72 & 49.98 & 49.78 & 45.01 & 54.21 & 66.85 & 62.83 & 58.45 \\
 & \textbf{hi} & 61.44 & 51.48 & 54.69 & 52.42 & 63.93 & 62.97 & 62.61 & 62.12 & 56.17 & 50.32 & 44.63 & 40.50 & 58.10 & 62.18 & 61.88 & 56.36 \\
 & \textbf{ru} & 46.27 & 55.03 & 54.31 & 53.23 & 49.98 & 53.01 & 53.23 & 56.03 & 56.15 & 46.57 & 49.84 & 43.17 & 50.64 & 53.97 & 53.31 & 51.65 \\
 & \textbf{sw} & 48.26 & 51.72 & 53.09 & 46.99 & 57.01 & 60.56 & 59.72 & 50.02 & 51.32 & 52.20 & 43.31 & 44.15 & 45.85 & 57.54 & 55.77 & 51.83 \\
 & \textbf{th} & 34.61 & 42.71 & 38.74 & 44.43 & 42.00 & 41.86 & 38.72 & 35.95 & 43.07 & 38.36 & 54.05 & 35.71 & 34.31 & 38.70 & 34.29 & 39.83 \\
 & \textbf{tr} & 37.68 & 45.69 & 44.97 & 44.19 & 42.28 & 41.94 & 38.90 & 39.42 & 44.57 & 39.54 & 40.52 & 51.66 & 34.19 & 37.58 & 37.84 & 41.40 \\
 & \textbf{ur} & 47.84 & 41.88 & 45.55 & 40.60 & 47.80 & 49.82 & 51.42 & 53.89 & 46.69 & 41.20 & 35.01 & 35.87 & 51.66 & 48.64 & 50.80 & 45.91 \\
 & \textbf{vi} & 62.04 & 52.53 & 53.39 & 47.84 & 67.64 & 66.29 & 66.51 & 58.90 & 56.71 & 50.42 & 44.53 & 37.90 & 52.48 & 66.51 & 63.51 & 56.48 \\
 & \textbf{zh} & 53.51 & 43.69 & 47.58 & 42.16 & 54.63 & 53.55 & 54.75 & 51.16 & 48.20 & 40.56 & 35.57 & 35.33 & 45.87 & 55.19 & 52.30 & 47.60 \\
 \midrule
\multirow{17}{*}{\textbf{BLOOM-1.7B}} & \textbf{Zero-shot} & 33.21 & 33.63 & 33.35 & 34.05 & 33.45 & 33.21 & 33.27 & 33.41 & 32.87 & 33.27 & 33.35 & 33.33 & 33.21 & 33.25 & 32.91 & 33.32 \\
 & \textbf{Average} & 59.07 & 53.05 & 54.84 & 52.24 & 64.00 & 62.63 & 62.97 & 57.94 & 55.41 & 52.07 & 48.42 & 44.14 & 54.54 & 60.36 & 58.89 & 56.04 \\
 & \textbf{ar} & 70.42 & 55.65 & 61.78 & 55.31 & 76.37 & 75.09 & 75.85 & 67.43 & 61.50 & 59.88 & 48.44 & 47.58 & 61.96 & 72.77 & 68.42 & 63.90 \\
 & \textbf{bg} & 50.06 & 60.08 & 57.21 & 56.83 & 53.53 & 52.95 & 52.63 & 50.34 & 59.36 & 47.27 & 54.13 & 42.12 & 50.46 & 50.54 & 52.20 & 52.65 \\
 & \textbf{de} & 61.06 & 56.43 & 62.85 & 57.90 & 69.64 & 66.55 & 67.84 & 58.50 & 59.52 & 55.65 & 52.26 & 47.11 & 55.05 & 62.10 & 60.64 & 59.54 \\
 & \textbf{el} & 47.33 & 54.97 & 52.16 & 58.74 & 50.80 & 49.58 & 49.86 & 46.45 & 56.25 & 45.23 & 53.21 & 39.20 & 46.17 & 48.46 & 45.53 & 49.60 \\
 & \textbf{en} & 62.04 & 48.56 & 52.50 & 46.15 & 73.49 & 68.96 & 69.56 & 59.36 & 53.87 & 54.17 & 45.33 & 40.92 & 55.69 & 63.83 & 60.98 & 57.03 \\
 & \textbf{es} & 69.12 & 50.18 & 56.31 & 46.01 & 76.93 & 76.37 & 75.19 & 64.63 & 51.46 & 57.41 & 41.60 & 44.77 & 60.96 & 71.06 & 70.34 & 60.82 \\
 & \textbf{fr} & 70.84 & 55.29 & 60.54 & 51.08 & 77.45 & 74.99 & 75.25 & 66.37 & 58.36 & 58.10 & 37.94 & 46.87 & 59.64 & 72.20 & 66.13 & 62.07 \\
 & \textbf{hi} & 64.71 & 55.15 & 53.63 & 55.33 & 63.31 & 65.81 & 66.43 & 65.51 & 56.61 & 55.09 & 48.30 & 43.49 & 59.76 & 64.69 & 67.01 & 58.99 \\
 & \textbf{ru} & 61.92 & 58.70 & 60.68 & 58.36 & 69.20 & 67.68 & 67.84 & 61.70 & 62.67 & 56.65 & 52.18 & 47.11 & 57.19 & 63.57 & 63.47 & 60.59 \\
 & \textbf{sw} & 61.22 & 57.11 & 58.28 & 55.73 & 67.39 & 64.49 & 65.95 & 60.36 & 58.82 & 57.09 & 51.84 & 46.27 & 55.87 & 63.67 & 60.38 & 58.96 \\
 & \textbf{th} & 40.28 & 43.45 & 41.34 & 45.45 & 41.90 & 42.46 & 41.68 & 42.93 & 41.38 & 38.24 & 54.13 & 35.59 & 40.30 & 41.74 & 38.52 & 41.96 \\
 & \textbf{tr} & 45.01 & 48.68 & 48.08 & 46.45 & 49.04 & 46.03 & 47.03 & 43.37 & 46.17 & 43.05 & 46.35 & 55.47 & 42.77 & 44.85 & 43.95 & 46.42 \\
 & \textbf{ur} & 50.98 & 44.11 & 48.10 & 44.69 & 51.50 & 51.04 & 52.16 & 56.05 & 48.68 & 43.67 & 42.50 & 36.09 & 55.55 & 51.00 & 54.01 & 48.68 \\
 & \textbf{vi} & 68.66 & 56.83 & 59.62 & 55.73 & 76.21 & 74.29 & 73.57 & 65.15 & 60.76 & 58.16 & 50.00 & 46.53 & 60.66 & 71.86 & 68.86 & 63.13 \\
 & \textbf{zh} & 62.34 & 50.60 & 49.56 & 49.90 & 63.29 & 63.23 & 63.71 & 60.90 & 55.73 & 51.42 & 48.06 & 42.99 & 56.03 & 63.03 & 62.91 & 56.25 \\
 \midrule
\multirow{17}{*}{\textbf{BLOOM-7.1B}} & \textbf{Zero-shot} & 33.15 & 33.29 & 32.57 & 33.33 & 33.33 & 33.55 & 32.79 & 33.13 & 33.21 & 33.33 & 33.31 & 33.33 & 33.65 & 33.31 & 33.33 & 33.24 \\
 & \textbf{Average} & 68.48 & 58.18 & 63.16 & 57.09 & 73.95 & 72.49 & 71.96 & 66.16 & 63.34 & 59.95 & 53.14 & 49.46 & 61.97 & 69.20 & 69.23 & 63.85 \\
 & \textbf{ar} & 75.63 & 61.14 & 67.27 & 58.78 & 82.67 & 80.78 & 79.28 & 72.63 & 67.80 & 64.23 & 54.37 & 51.82 & 68.12 & 76.15 & 75.69 & 69.09 \\
 & \textbf{bg} & 66.67 & 64.97 & 66.09 & 60.34 & 70.98 & 68.10 & 68.04 & 66.61 & 66.95 & 58.56 & 56.69 & 52.48 & 63.17 & 66.41 & 65.77 & 64.12 \\
 & \textbf{de} & 72.53 & 61.18 & 71.32 & 60.50 & 81.12 & 77.47 & 77.01 & 68.94 & 67.19 & 65.21 & 53.27 & 51.58 & 64.83 & 73.23 & 73.53 & 67.93 \\
 & \textbf{el} & 56.83 & 61.58 & 64.13 & 62.14 & 63.45 & 59.44 & 61.44 & 59.54 & 63.59 & 52.50 & 56.57 & 51.10 & 56.87 & 59.92 & 61.02 & 59.34 \\
 & \textbf{en} & 69.46 & 50.52 & 59.72 & 51.22 & 81.38 & 77.96 & 75.25 & 62.75 & 59.60 & 56.25 & 46.59 & 44.15 & 57.33 & 69.96 & 70.72 & 62.19 \\
 & \textbf{es} & 76.61 & 56.61 & 64.21 & 55.51 & 83.95 & 80.98 & 80.20 & 72.32 & 65.43 & 64.03 & 51.92 & 48.32 & 66.87 & 76.01 & 76.63 & 67.97 \\
 & \textbf{fr} & 75.77 & 55.55 & 62.71 & 54.59 & 82.20 & 80.76 & 80.00 & 71.96 & 64.03 & 63.29 & 50.58 & 46.75 & 66.57 & 76.13 & 76.63 & 67.17 \\
 & \textbf{hi} & 75.05 & 61.84 & 67.52 & 61.16 & 80.14 & 78.56 & 77.58 & 73.09 & 68.46 & 65.39 & 54.87 & 51.58 & 68.92 & 75.29 & 75.83 & 69.02 \\
 & \textbf{ru} & 73.29 & 63.23 & 69.68 & 58.78 & 80.26 & 77.33 & 77.09 & 69.10 & 68.76 & 65.65 & 54.57 & 51.58 & 66.07 & 73.85 & 73.39 & 68.18 \\
 & \textbf{sw} & 72.22 & 61.80 & 67.27 & 58.04 & 77.39 & 76.43 & 76.01 & 68.74 & 65.99 & 64.73 & 55.27 & 52.02 & 64.21 & 72.83 & 72.73 & 67.05 \\
 & \textbf{th} & 48.10 & 49.98 & 48.04 & 52.77 & 53.85 & 51.20 & 51.14 & 51.06 & 48.90 & 45.13 & 56.83 & 40.58 & 46.85 & 49.56 & 50.96 & 49.66 \\
 & \textbf{tr} & 54.33 & 54.57 & 55.69 & 52.81 & 53.97 & 57.01 & 56.05 & 50.62 & 53.97 & 53.35 & 50.94 & 56.85 & 46.91 & 55.59 & 50.44 & 53.54 \\
 & \textbf{ur} & 62.99 & 50.22 & 51.96 & 51.58 & 55.95 & 64.89 & 65.41 & 63.35 & 56.39 & 51.94 & 47.05 & 42.12 & 60.40 & 65.03 & 64.49 & 56.92 \\
 & \textbf{vi} & 75.97 & 59.98 & 65.91 & 58.52 & 82.89 & 80.58 & 80.10 & 72.36 & 67.31 & 65.67 & 53.01 & 51.02 & 67.68 & 75.63 & 76.33 & 68.86 \\
 & \textbf{zh} & 71.76 & 59.54 & 65.91 & 59.66 & 79.06 & 75.85 & 74.79 & 69.26 & 65.69 & 63.39 & 54.55 & 50.02 & 64.79 & 72.44 & 74.29 & 66.73 \\
 \bottomrule
\end{tabular}}
\end{table}

\begin{table}[ht]
\caption{Detailed results of Pythia on XNLI dataset.}
\centering
\label{tab:pythia-xnli}
\scalebox{0.75}{
\begin{tabular}{llcccccccccccccccc}
\toprule
\multirow{2}{*}{\textbf{Model}} & \multicolumn{1}{c}{\multirow{2}{*}{\textbf{Train}}} & \multicolumn{15}{c}{\textbf{Test}} & \multicolumn{1}{c}{\multirow{2}{*}{\textbf{Average}}} \\
 & \multicolumn{1}{c}{} & \textbf{ar} & \textbf{bg} & \textbf{de} & \textbf{el} & \textbf{en} & \textbf{es} & \textbf{fr} & \textbf{hi} & \textbf{ru} & \textbf{sw} & \textbf{th} & \textbf{tr} & \textbf{ur} & \textbf{vi} & \textbf{zh} & \multicolumn{1}{c}{} \\
 \hline
\multirow{17}{*}{\textbf{Pythia-410m}} & \textbf{Zero-shot} & 33.27 & 32.24 & 32.93 & 33.87 & 32.71 & 33.43 & 33.59 & 33.43 & 32.46 & 33.01 & 33.43 & 33.29 & 33.05 & 32.63 & 33.39 & 33.12 \\
 & \textbf{Average} & 46.21 & 47.57 & 47.94 & 47.68 & 49.94 & 48.46 & 48.80 & 42.27 & 46.98 & 40.19 & 45.58 & 41.28 & 40.49 & 47.15 & 47.55 & 45.87 \\
 & \textbf{ar} & 50.64 & 45.31 & 44.67 & 44.35 & 42.99 & 43.35 & 43.93 & 41.08 & 45.15 & 42.26 & 43.27 & 42.55 & 38.94 & 44.91 & 44.61 & 43.87 \\
 & \textbf{bg} & 47.29 & 51.86 & 49.04 & 49.36 & 45.57 & 48.44 & 48.48 & 40.82 & 50.24 & 41.10 & 43.93 & 42.26 & 37.76 & 48.98 & 49.72 & 46.32 \\
 & \textbf{de} & 46.53 & 53.49 & 59.56 & 49.50 & 63.39 & 59.60 & 58.96 & 41.48 & 50.82 & 45.11 & 47.31 & 45.07 & 38.50 & 51.08 & 53.31 & 50.91 \\
 & \textbf{el} & 48.24 & 48.88 & 50.98 & 54.17 & 44.29 & 51.60 & 51.98 & 46.43 & 49.06 & 41.68 & 48.12 & 44.35 & 43.63 & 51.16 & 46.99 & 48.10 \\
 & \textbf{en} & 49.26 & 52.77 & 55.69 & 52.93 & 74.13 & 60.94 & 60.80 & 45.07 & 52.97 & 43.47 & 47.50 & 43.17 & 43.59 & 50.74 & 52.48 & 52.37 \\
 & \textbf{es} & 53.81 & 57.01 & 60.60 & 55.21 & 69.28 & 63.71 & 63.95 & 49.90 & 56.33 & 40.84 & 52.30 & 46.97 & 46.59 & 55.11 & 58.66 & 55.35 \\
 & \textbf{fr} & 52.81 & 55.93 & 58.20 & 53.63 & 65.63 & 61.12 & 61.54 & 48.20 & 54.81 & 44.25 & 52.89 & 45.67 & 44.27 & 54.01 & 55.97 & 53.93 \\
 & \textbf{hi} & 42.36 & 41.96 & 39.36 & 45.47 & 39.58 & 40.80 & 40.86 & 49.20 & 42.97 & 38.54 & 46.39 & 38.54 & 43.87 & 44.89 & 41.88 & 42.44 \\
 & \textbf{ru} & 49.56 & 56.63 & 55.91 & 53.69 & 55.83 & 56.59 & 56.49 & 43.61 & 55.47 & 41.30 & 47.60 & 44.95 & 39.48 & 51.40 & 55.63 & 50.94 \\
 & \textbf{sw} & 38.86 & 35.73 & 37.43 & 38.56 & 37.50 & 37.56 & 37.70 & 34.69 & 36.67 & 44.57 & 35.37 & 38.72 & 35.61 & 38.38 & 36.25 & 37.57 \\
 & \textbf{th} & 42.06 & 37.45 & 37.80 & 41.92 & 39.12 & 39.84 & 38.10 & 38.76 & 36.55 & 33.57 & 49.64 & 34.51 & 38.08 & 43.53 & 42.40 & 39.56 \\
 & \textbf{tr} & 44.33 & 45.13 & 43.71 & 45.17 & 44.83 & 43.51 & 43.95 & 41.28 & 43.27 & 44.09 & 43.71 & 47.92 & 40.38 & 42.99 & 43.41 & 43.85 \\
 & \textbf{ur} & 35.21 & 36.95 & 35.75 & 35.99 & 35.09 & 33.95 & 34.99 & 34.93 & 35.75 & 33.21 & 35.91 & 33.61 & 40.80 & 36.17 & 36.71 & 35.67 \\
 & \textbf{vi} & 46.15 & 45.39 & 41.40 & 48.38 & 41.86 & 39.42 & 41.98 & 37.80 & 44.47 & 33.77 & 44.37 & 36.25 & 36.59 & 48.42 & 42.44 & 41.91 \\
 & \textbf{zh} & 46.07 & 49.04 & 49.06 & 46.93 & 50.06 & 46.49 & 48.32 & 40.74 & 50.18 & 35.07 & 45.37 & 34.71 & 39.20 & 45.45 & 52.77 & 45.30 \\
 \hline
\multirow{17}{*}{\textbf{Pythia-1.4b}} & \textbf{Zero-shot} & 34.15 & 33.35 & 33.37 & 33.21 & 33.53 & 33.35 & 33.49 & 33.41 & 33.33 & 33.33 & 33.33 & 33.31 & 33.27 & 33.33 & 33.33 & 33.41 \\
 & \textbf{Average} & 52.96 & 57.12 & 58.93 & 55.67 & 63.49 & 60.73 & 60.37 & 50.34 & 55.73 & 45.06 & 52.86 & 53.09 & 49.49 & 55.33 & 56.51 & 55.18 \\
 & \textbf{ar} & 59.50 & 62.04 & 62.04 & 61.54 & 62.02 & 63.19 & 60.58 & 53.41 & 59.86 & 44.81 & 56.03 & 57.01 & 51.62 & 60.52 & 61.48 & 58.38 \\
 & \textbf{bg} & 53.67 & 61.60 & 62.87 & 56.09 & 67.39 & 63.73 & 64.21 & 49.56 & 58.86 & 47.05 & 54.17 & 53.01 & 49.08 & 56.67 & 61.80 & 57.32 \\
 & \textbf{de} & 54.61 & 61.54 & 67.13 & 57.31 & 76.13 & 68.18 & 68.62 & 50.14 & 59.92 & 46.95 & 53.81 & 57.45 & 49.24 & 57.19 & 62.73 & 59.40 \\
 & \textbf{el} & 57.31 & 62.02 & 66.05 & 62.04 & 71.02 & 66.97 & 66.83 & 53.87 & 60.78 & 48.12 & 56.47 & 58.40 & 52.83 & 59.82 & 63.11 & 60.38 \\
 & \textbf{en} & 45.67 & 52.26 & 58.50 & 50.28 & 78.66 & 64.07 & 63.55 & 44.29 & 50.96 & 44.67 & 49.84 & 50.14 & 43.45 & 52.14 & 54.97 & 53.56 \\
 & \textbf{es} & 54.93 & 62.26 & 66.75 & 60.08 & 78.56 & 71.12 & 69.82 & 48.32 & 61.66 & 46.45 & 55.37 & 54.79 & 49.70 & 60.16 & 62.91 & 60.19 \\
 & \textbf{fr} & 56.45 & 62.99 & 66.77 & 60.94 & 77.29 & 70.32 & 70.14 & 50.66 & 62.20 & 46.87 & 54.97 & 56.91 & 48.76 & 59.18 & 59.06 & 60.23 \\
 & \textbf{hi} & 51.18 & 50.38 & 48.46 & 51.48 & 48.96 & 51.58 & 50.22 & 54.91 & 49.96 & 40.14 & 49.74 & 47.03 & 50.18 & 52.71 & 46.97 & 49.59 \\
 & \textbf{ru} & 55.65 & 61.56 & 63.47 & 60.74 & 70.16 & 64.77 & 64.47 & 50.04 & 61.10 & 44.33 & 55.03 & 54.63 & 48.88 & 58.44 & 59.30 & 58.17 \\
 & \textbf{sw} & 52.50 & 52.95 & 51.28 & 53.69 & 49.60 & 51.72 & 51.02 & 49.90 & 51.24 & 50.46 & 48.58 & 52.26 & 48.16 & 50.78 & 44.77 & 50.59 \\
 & \textbf{th} & 53.21 & 53.45 & 51.20 & 51.98 & 50.08 & 52.95 & 53.43 & 50.26 & 51.70 & 43.95 & 56.59 & 49.12 & 49.78 & 53.29 & 56.51 & 51.83 \\
 & \textbf{tr} & 51.66 & 52.81 & 55.97 & 54.59 & 50.98 & 56.79 & 56.31 & 53.79 & 52.77 & 44.83 & 52.16 & 58.10 & 49.22 & 52.50 & 49.04 & 52.77 \\
 & \textbf{ur} & 35.75 & 36.99 & 36.27 & 36.15 & 34.05 & 35.11 & 36.97 & 38.64 & 36.49 & 33.63 & 37.84 & 35.17 & 46.91 & 37.33 & 40.18 & 37.16 \\
 & \textbf{vi} & 56.17 & 61.28 & 63.09 & 58.88 & 67.05 & 63.13 & 62.87 & 54.61 & 58.22 & 45.87 & 56.37 & 56.87 & 51.66 & 61.42 & 59.58 & 58.47 \\
 & \textbf{zh} & 56.17 & 62.63 & 64.13 & 59.24 & 70.48 & 67.25 & 66.57 & 52.65 & 60.24 & 47.70 & 55.93 & 55.47 & 52.93 & 57.76 & 65.29 & 59.63 \\
 \hline
\multirow{17}{*}{\textbf{Pythia-6.9b}} & \textbf{Zero-shot} & 33.33 & 33.33 & 33.33 & 33.33 & 33.33 & 33.21 & 33.33 & 33.33 & 33.33 & 33.33 & 33.33 & 33.33 & 33.33 & 33.33 & 33.33 & 33.33 \\
 & \textbf{Average} & 61.61 & 65.37 & 67.14 & 65.38 & 72.44 & 69.35 & 69.23 & 54.97 & 64.61 & 49.66 & 57.88 & 60.41 & 54.36 & 62.42 & 64.92 & 62.65 \\
 & \textbf{ar} & 65.47 & 68.66 & 70.18 & 68.84 & 72.18 & 70.66 & 71.56 & 55.85 & 68.32 & 52.14 & 60.18 & 62.73 & 55.07 & 66.47 & 68.66 & 65.13 \\
 & \textbf{bg} & 63.33 & 69.06 & 70.66 & 67.92 & 76.05 & 73.93 & 73.35 & 55.71 & 67.49 & 51.42 & 60.02 & 63.19 & 53.89 & 65.09 & 68.82 & 65.33 \\
 & \textbf{de} & 64.19 & 69.30 & 73.43 & 69.30 & 80.72 & 76.75 & 76.25 & 55.87 & 69.30 & 51.62 & 61.26 & 63.91 & 56.59 & 65.97 & 68.56 & 66.87 \\
 & \textbf{el} & 63.67 & 70.42 & 71.58 & 68.52 & 77.19 & 74.95 & 73.73 & 56.41 & 68.28 & 52.20 & 58.92 & 63.67 & 55.07 & 63.63 & 68.60 & 65.79 \\
 & \textbf{en} & 56.03 & 61.88 & 65.89 & 61.50 & 83.77 & 70.84 & 70.10 & 47.31 & 61.10 & 45.59 & 51.42 & 54.39 & 46.31 & 55.91 & 61.84 & 59.59 \\
 & \textbf{es} & 64.41 & 68.98 & 73.05 & 68.74 & 83.41 & 76.77 & 76.15 & 55.29 & 68.42 & 50.20 & 57.80 & 61.24 & 53.55 & 64.57 & 68.50 & 66.07 \\
 & \textbf{fr} & 62.85 & 68.50 & 72.12 & 68.80 & 81.36 & 77.68 & 75.89 & 54.31 & 68.34 & 49.92 & 57.25 & 62.59 & 53.83 & 62.71 & 69.54 & 65.71 \\
 & \textbf{hi} & 60.00 & 60.54 & 59.40 & 62.51 & 60.00 & 57.84 & 59.64 & 60.08 & 59.26 & 48.00 & 57.41 & 59.18 & 56.25 & 61.02 & 57.68 & 58.59 \\
 & \textbf{ru} & 62.55 & 68.52 & 70.74 & 68.06 & 79.18 & 75.25 & 73.99 & 54.79 & 69.22 & 50.52 & 57.31 & 61.82 & 54.77 & 64.53 & 68.18 & 65.30 \\
 & \textbf{sw} & 58.48 & 59.36 & 61.10 & 60.22 & 60.96 & 61.00 & 61.50 & 53.03 & 57.35 & 54.25 & 56.31 & 58.34 & 53.03 & 60.70 & 58.30 & 58.26 \\
 & \textbf{th} & 62.38 & 62.73 & 63.87 & 65.13 & 68.80 & 62.67 & 64.07 & 58.46 & 62.57 & 50.28 & 62.50 & 59.96 & 56.37 & 64.27 & 64.67 & 61.92 \\
 & \textbf{tr} & 62.18 & 65.73 & 69.08 & 65.89 & 72.55 & 70.00 & 70.62 & 56.05 & 65.15 & 50.92 & 58.04 & 63.41 & 55.23 & 63.43 & 65.17 & 63.56 \\
 & \textbf{ur} & 50.14 & 47.56 & 43.49 & 46.77 & 37.84 & 44.05 & 44.35 & 48.22 & 46.37 & 35.05 & 48.50 & 45.07 & 53.21 & 44.29 & 47.37 & 45.49 \\
 & \textbf{vi} & 63.33 & 68.76 & 69.96 & 68.40 & 71.96 & 72.26 & 72.34 & 54.65 & 67.58 & 52.18 & 60.30 & 63.01 & 55.47 & 66.67 & 66.47 & 64.89 \\
 & \textbf{zh} & 65.13 & 70.48 & 72.57 & 70.08 & 80.56 & 75.61 & 74.97 & 58.48 & 70.40 & 50.66 & 61.04 & 63.67 & 56.81 & 67.11 & 71.50 & 67.27 \\
 \bottomrule
\end{tabular}}
\end{table}

\begin{table}[ht]
\caption{Detailed results of LLaMA on XNLI dataset.}
\centering
\label{tab:llama-xnli}
\scalebox{0.75}{
\begin{tabular}{llccccccccccccccccc}
\toprule
\multirow{2}{*}{\textbf{Model}} & \multicolumn{1}{c}{\multirow{2}{*}{\textbf{Train}}} & \multicolumn{15}{c}{\textbf{Test}} & \multicolumn{1}{c}{\multirow{2}{*}{\textbf{Average}}} \\
 & \multicolumn{1}{c}{} & \textbf{ar} & \textbf{bg} & \textbf{de} & \textbf{el} & \textbf{en} & \textbf{es} & \textbf{fr} & \textbf{hi} & \textbf{ru} & \textbf{sw} & \textbf{th} & \textbf{tr} & \textbf{ur} & \textbf{vi} & \textbf{zh} & \multicolumn{1}{c}{} \\
 \midrule
\multirow{17}{*}{\textbf{LLaMA-6.7B}} & \textbf{Zero-shot} & 33.47 & 33.83 & 33.45 & 33.35 & 33.75 & 34.47 & 34.45 & 33.65 & 33.85 & 33.19 & 33.33 & 33.31 & 33.49 & 33.33 & 33.41 & 33.62 \\
 & \textbf{Average} & 58.32 & 70.74 & 70.68 & 60.90 & 75.32 & 72.25 & 71.99 & 55.28 & 70.04 & 45.79 & 52.19 & 54.42 & 50.98 & 56.71 & 65.52 & 62.07 \\
 & \textbf{ar} & 64.75 & 76.55 & 76.75 & 65.27 & 82.83 & 78.64 & 78.04 & 58.96 & 75.47 & 49.36 & 57.43 & 59.04 & 54.51 & 61.30 & 71.60 & 67.37 \\
 & \textbf{bg} & 61.22 & 79.36 & 79.86 & 64.93 & 85.53 & 81.94 & 80.92 & 56.33 & 77.54 & 46.69 & 52.59 & 57.37 & 50.98 & 60.00 & 71.14 & 67.09 \\
 & \textbf{de} & 61.04 & 79.32 & 80.92 & 64.99 & 87.78 & 82.99 & 81.96 & 54.89 & 78.24 & 46.11 & 50.60 & 56.53 & 48.82 & 58.50 & 70.38 & 66.87 \\
 & \textbf{el} & 64.03 & 76.79 & 77.78 & 68.00 & 80.70 & 78.76 & 78.18 & 62.08 & 76.43 & 49.84 & 54.99 & 58.92 & 56.23 & 61.90 & 71.66 & 67.75 \\
 & \textbf{en} & 52.69 & 72.79 & 75.81 & 57.39 & 86.85 & 77.56 & 76.99 & 46.97 & 73.09 & 40.58 & 46.57 & 51.12 & 45.51 & 54.71 & 61.82 & 61.36 \\
 & \textbf{es} & 55.99 & 76.57 & 79.22 & 60.66 & 86.55 & 81.00 & 81.22 & 52.89 & 76.31 & 42.63 & 46.59 & 53.91 & 46.63 & 56.09 & 67.62 & 64.26 \\
 & \textbf{fr} & 58.20 & 76.73 & 79.88 & 60.20 & 86.59 & 81.38 & 80.04 & 52.91 & 77.05 & 44.05 & 49.56 & 54.71 & 48.38 & 57.60 & 68.56 & 65.06 \\
 & \textbf{hi} & 61.82 & 70.58 & 71.32 & 66.17 & 73.81 & 72.18 & 72.79 & 63.23 & 71.50 & 47.82 & 56.31 & 57.21 & 57.41 & 59.08 & 68.78 & 64.67 \\
 & \textbf{ru} & 61.54 & 77.78 & 79.42 & 64.35 & 85.87 & 82.10 & 80.66 & 57.98 & 77.96 & 47.68 & 52.26 & 56.75 & 51.46 & 59.36 & 70.06 & 67.02 \\
 & \textbf{sw} & 53.33 & 53.47 & 47.35 & 52.67 & 52.20 & 49.52 & 49.32 & 50.08 & 52.30 & 52.57 & 50.70 & 47.19 & 48.12 & 48.46 & 54.53 & 50.79 \\
 & \textbf{th} & 50.40 & 53.77 & 50.04 & 51.30 & 50.52 & 49.72 & 51.98 & 48.32 & 52.20 & 45.29 & 57.41 & 45.99 & 46.65 & 48.38 & 56.83 & 50.59 \\
 & \textbf{tr} & 60.50 & 72.10 & 72.22 & 64.35 & 71.90 & 73.55 & 73.01 & 57.94 & 70.74 & 47.86 & 54.81 & 61.12 & 52.48 & 59.84 & 64.83 & 63.82 \\
 & \textbf{ur} & 50.28 & 48.98 & 41.72 & 50.08 & 43.61 & 44.25 & 44.65 & 52.69 & 47.27 & 35.57 & 47.92 & 42.28 & 53.31 & 45.47 & 47.84 & 46.39 \\
 & \textbf{vi} & 60.84 & 73.11 & 74.01 & 63.09 & 75.77 & 74.31 & 74.63 & 57.41 & 72.69 & 47.60 & 52.42 & 58.10 & 50.48 & 61.78 & 67.37 & 64.24 \\
 & \textbf{zh} & 58.20 & 73.19 & 73.95 & 59.98 & 79.24 & 75.79 & 75.41 & 56.45 & 71.76 & 43.17 & 52.63 & 56.01 & 53.79 & 58.16 & 69.84 & 63.84 \\
 \midrule
\multirow{17}{*}{\textbf{LLaMA-13B}} & \textbf{Zero-shot} & 32.95 & 33.39 & 33.35 & 33.33 & 33.95 & 33.49 & 33.23 & 33.31 & 33.11 & 33.23 & 33.37 & 33.35 & 33.29 & 33.37 & 33.17 & 33.33 \\
 & \textbf{Average} & 62.07 & 75.50 & 76.25 & 63.31 & 80.75 & 77.85 & 77.52 & 60.64 & 74.30 & 47.24 & 56.55 & 60.73 & 55.51 & 60.57 & 70.81 & 66.64 \\
 & \textbf{ar} & 66.89 & 78.70 & 79.26 & 66.77 & 83.93 & 81.50 & 81.00 & 64.53 & 77.03 & 49.88 & 59.96 & 63.79 & 58.56 & 63.65 & 74.69 & 70.01 \\
 & \textbf{bg} & 64.77 & 80.66 & 81.36 & 67.64 & 86.91 & 83.47 & 82.93 & 62.63 & 79.58 & 48.74 & 58.72 & 63.27 & 56.91 & 63.29 & 73.95 & 70.32 \\
 & \textbf{de} & 64.65 & 80.46 & 82.51 & 66.81 & 88.68 & 84.33 & 83.65 & 60.22 & 79.06 & 48.62 & 56.77 & 63.15 & 54.11 & 62.00 & 73.53 & 69.90 \\
 & \textbf{el} & 65.93 & 79.26 & 80.36 & 69.24 & 85.89 & 82.71 & 81.36 & 64.69 & 78.80 & 50.22 & 59.20 & 64.47 & 58.66 & 64.43 & 74.31 & 70.64 \\
 & \textbf{en} & 55.65 & 73.37 & 76.97 & 56.71 & 87.88 & 78.04 & 77.90 & 54.41 & 73.73 & 43.25 & 52.22 & 55.57 & 48.70 & 56.75 & 68.32 & 63.96 \\
 & \textbf{es} & 62.42 & 79.58 & 80.96 & 64.89 & 87.19 & 82.93 & 82.20 & 59.42 & 77.50 & 44.41 & 53.91 & 60.28 & 52.30 & 59.78 & 72.00 & 67.98 \\
 & \textbf{fr} & 62.57 & 80.52 & 82.18 & 64.19 & 89.66 & 83.99 & 83.03 & 60.48 & 79.52 & 46.13 & 55.97 & 61.16 & 54.23 & 61.28 & 73.95 & 69.26 \\
 & \textbf{hi} & 65.09 & 75.69 & 75.55 & 67.64 & 79.26 & 77.96 & 78.30 & 66.45 & 74.87 & 47.94 & 59.68 & 62.69 & 60.06 & 62.83 & 73.51 & 68.50 \\
 & \textbf{ru} & 63.65 & 80.44 & 81.74 & 66.71 & 88.12 & 83.43 & 82.48 & 60.96 & 78.80 & 46.77 & 56.77 & 64.11 & 55.43 & 62.10 & 73.91 & 69.69 \\
 & \textbf{sw} & 58.66 & 64.87 & 63.83 & 60.32 & 64.27 & 64.79 & 66.55 & 56.05 & 63.13 & 53.51 & 56.51 & 57.47 & 53.51 & 57.41 & 61.60 & 60.17 \\
 & \textbf{th} & 62.77 & 73.65 & 72.85 & 63.31 & 73.09 & 75.03 & 74.81 & 60.68 & 72.89 & 47.56 & 60.84 & 60.80 & 56.25 & 61.74 & 71.32 & 65.84 \\
 & \textbf{tr} & 63.65 & 76.69 & 77.80 & 66.25 & 84.05 & 80.18 & 79.14 & 62.24 & 74.25 & 49.76 & 56.71 & 64.79 & 57.90 & 62.14 & 70.74 & 68.42 \\
 & \textbf{ur} & 49.62 & 53.99 & 53.69 & 42.36 & 48.08 & 53.61 & 53.25 & 56.31 & 54.23 & 38.52 & 43.37 & 46.85 & 55.17 & 46.55 & 54.05 & 49.98 \\
 & \textbf{vi} & 64.35 & 77.52 & 78.08 & 67.37 & 81.44 & 80.62 & 80.04 & 62.06 & 76.21 & 50.18 & 60.06 & 63.01 & 57.33 & 64.75 & 73.13 & 69.08 \\
 & \textbf{zh} & 60.38 & 77.03 & 76.67 & 59.40 & 82.79 & 75.09 & 76.23 & 58.54 & 74.91 & 43.15 & 57.49 & 59.54 & 53.51 & 59.78 & 73.19 & 65.85 \\
 \midrule
\multirow{17}{*}{\textbf{LLaMA-32.5B}} & \textbf{Zero-shot} & 33.03 & 34.01 & 33.31 & 33.01 & 38.92 & 33.83 & 33.95 & 33.69 & 33.53 & 32.46 & 31.92 & 32.30 & 33.19 & 34.29 & 33.57 & 33.67 \\
 & \textbf{Average} & 68.63 & 80.68 & 81.27 & 72.29 & 86.00 & 83.36 & 82.51 & 66.20 & 79.37 & 50.63 & 60.15 & 65.88 & 59.56 & 67.47 & 76.65 & 72.04 \\
 & \textbf{ar} & 71.48 & 82.32 & 81.78 & 75.09 & 84.87 & 84.39 & 83.93 & 68.94 & 80.26 & 52.67 & 63.39 & 67.58 & 62.24 & 70.58 & 78.46 & 73.87 \\
 & \textbf{bg} & 70.84 & 83.27 & 84.57 & 74.45 & 89.92 & 86.59 & 85.93 & 68.46 & 82.30 & 51.82 & 60.50 & 67.07 & 61.12 & 69.82 & 78.84 & 74.37 \\
 & \textbf{de} & 67.94 & 82.53 & 83.49 & 72.22 & 90.18 & 85.97 & 85.07 & 65.05 & 81.20 & 49.96 & 58.54 & 65.79 & 57.17 & 67.49 & 77.05 & 72.64 \\
 & \textbf{el} & 73.07 & 82.63 & 82.63 & 76.87 & 86.53 & 85.29 & 84.47 & 69.58 & 80.90 & 50.52 & 65.27 & 68.58 & 63.35 & 71.28 & 79.18 & 74.68 \\
 & \textbf{en} & 59.92 & 78.46 & 80.14 & 64.29 & 90.58 & 82.97 & 82.00 & 57.82 & 78.94 & 44.01 & 49.20 & 58.70 & 51.04 & 59.36 & 74.41 & 67.46 \\
 & \textbf{es} & 66.79 & 81.42 & 84.01 & 70.46 & 90.36 & 85.85 & 84.89 & 63.97 & 81.00 & 49.88 & 56.71 & 65.65 & 56.95 & 65.93 & 77.41 & 72.09 \\
 & \textbf{fr} & 66.15 & 81.62 & 83.93 & 70.84 & 90.02 & 85.73 & 84.53 & 63.31 & 80.42 & 51.16 & 58.86 & 65.43 & 55.77 & 64.29 & 77.49 & 71.97 \\
 & \textbf{hi} & 71.12 & 81.22 & 81.88 & 75.17 & 85.53 & 84.17 & 83.47 & 70.48 & 80.14 & 50.90 & 63.29 & 67.94 & 63.15 & 69.46 & 78.20 & 73.74 \\
 & \textbf{ru} & 69.58 & 83.03 & 84.39 & 72.79 & 90.08 & 87.03 & 85.77 & 66.05 & 82.12 & 51.70 & 59.86 & 66.97 & 59.96 & 68.86 & 78.34 & 73.77 \\
 & \textbf{sw} & 64.35 & 77.56 & 79.34 & 66.61 & 84.25 & 81.12 & 80.20 & 62.67 & 76.37 & 57.15 & 58.46 & 64.77 & 56.53 & 66.13 & 74.03 & 69.97 \\
 & \textbf{th} & 71.20 & 81.24 & 80.56 & 75.03 & 82.16 & 82.34 & 81.64 & 68.16 & 78.86 & 50.32 & 64.19 & 66.61 & 62.16 & 69.90 & 78.08 & 72.83 \\
 & \textbf{tr} & 69.40 & 81.40 & 82.48 & 73.13 & 88.20 & 85.17 & 83.99 & 67.80 & 80.40 & 52.44 & 60.94 & 69.42 & 60.68 & 69.22 & 76.73 & 73.43 \\
 & \textbf{ur} & 65.31 & 68.62 & 65.47 & 67.72 & 66.99 & 65.07 & 64.83 & 64.25 & 66.71 & 43.15 & 57.19 & 58.46 & 59.06 & 59.72 & 64.39 & 62.46 \\
 & \textbf{vi} & 71.78 & 82.32 & 82.42 & 76.17 & 84.57 & 84.27 & 83.09 & 69.14 & 80.14 & 53.49 & 64.65 & 67.76 & 63.25 & 71.98 & 77.90 & 74.20 \\
 & \textbf{zh} & 70.48 & 82.53 & 82.00 & 73.57 & 85.83 & 84.41 & 83.87 & 67.29 & 80.78 & 50.28 & 61.18 & 67.49 & 60.92 & 68.04 & 79.24 & 73.19 \\
 \bottomrule
\end{tabular}}
\end{table}

\begin{table}[ht]
\caption{Detailed results of BLOOM, Pythia, and LLaMA on XCOPA, LogiQA, and GSM8K datasets.}
\centering
\label{tab:others}
\scalebox{0.8}{
\begin{tabular}{llccccccccccccccccc}
\toprule
\multirow{2}{*}{\textbf{Model}} & \multicolumn{1}{c}{\multirow{2}{*}{\textbf{Train}}} & \multicolumn{4}{c}{\textbf{XCOPA}} & \multicolumn{4}{c}{\textbf{LogiQA}} & \multicolumn{4}{c}{\textbf{GSM8K}} \\
\cmidrule(lr){3-6}
\cmidrule(lr){7-10}
\cmidrule(lr){11-14}
 & \multicolumn{1}{c}{} & \textbf{en} & \textbf{fr} & \textbf{zh} & \textbf{Average} & \textbf{en} & \textbf{fr} & \textbf{zh} & \textbf{Average} & \textbf{en} & \textbf{fr} & \textbf{zh} & \textbf{Average} \\
 \midrule
\multirow{5}{*}{BLOOM-560M} & Zero-shot & 50.00 & 50.00 & 50.00 & 50.00 & 20.28 & 20.28 & 20.28 & 20.28 & 2.40 & 2.00 & 1.20 & 1.87 \\
 & Average & 48.73 & 50.13 & 51.13 & 50.00 & 22.63 & 24.37 & 22.43 & 23.14 & 3.07 & 1.73 & 1.33 & 2.04 \\
 & en & 49.00 & 52.00 & 48.80 & 49.93 & 23.50 & 25.19 & 22.27 & 23.66 & 4.80 & 2.00 & 0.00 & 2.27 \\
 & fr & 48.80 & 51.40 & 51.40 & 50.53 & 23.20 & 25.35 & 20.43 & 22.99 & 2.00 & 1.60 & 0.80 & 1.47 \\
 & zh & 48.40 & 47.00 & 53.20 & 49.53 & 21.20 & 22.58 & 24.58 & 22.79 & 2.40 & 1.60 & 3.20 & 2.40 \\
 \midrule
\multirow{5}{*}{BLOOM-1.7B} & Zero-shot & 49.20 & 49.80 & 50.00 & 49.67 & 19.97 & 20.58 & 20.28 & 20.28 & 1.60 & 2.40 & 2.40 & 2.13 \\
 & Average & 49.13 & 50.20 & 51.00 & 50.11 & 25.14 & 26.11 & 22.32 & 24.53 & 4.40 & 4.27 & 4.27 & 4.31 \\
 & en & 48.20 & 50.20 & 50.20 & 49.53 & 25.65 & 25.35 & 21.04 & 24.01 & 5.60 & 5.20 & 4.00 & 4.93 \\
 & fr & 49.60 & 49.80 & 50.60 & 50.00 & 26.73 & 28.11 & 22.73 & 25.86 & 4.00 & 4.80 & 2.80 & 3.87 \\
 & zh & 49.60 & 50.60 & 52.20 & 50.80 & 23.04 & 24.88 & 23.20 & 23.71 & 3.60 & 2.80 & 6.00 & 4.13 \\
 \midrule
\multirow{5}{*}{BLOOM-7.1B} & Zero-shot & 49.80 & 51.60 & 50.00 & 50.47 & 23.04 & 20.89 & 19.97 & 21.30 & 2.80 & 3.20 & 2.40 & 2.80 \\
 & Average & 52.00 & 49.33 & 50.67 & 50.67 & 26.16 & 26.73 & 24.83 & 25.91 & 11.07 & 12.00 & 8.27 & 10.44 \\
 & en & 54.00 & 48.80 & 51.40 & 51.40 & 25.81 & 23.96 & 23.35 & 24.37 & 11.60 & 14.00 & 8.80 & 11.47 \\
 & fr & 50.20 & 49.00 & 50.40 & 49.87 & 26.27 & 28.73 & 25.81 & 26.93 & 11.20 & 10.00 & 7.60 & 9.60 \\
 & zh & 51.80 & 50.20 & 50.20 & 50.73 & 26.42 & 27.50 & 25.35 & 26.42 & 10.40 & 12.00 & 8.40 & 10.27 \\
 \midrule
\multirow{5}{*}{Pythia-410M} & Zero-shot & 50.00 & 49.80 & 50.00 & 49.93 & 20.28 & 23.50 & 20.28 & 21.35 & 2.80 & 2.00 & 3.20 & 2.67 \\
 & Average & 50.13 & 50.80 & 48.93 & 49.96 & 23.76 & 21.76 & 21.81 & 22.44 & 2.53 & 2.27 & 2.00 & 2.27 \\
 & en & 50.20 & 50.40 & 49.80 & 50.13 & 25.65 & 22.12 & 21.66 & 23.14 & 2.40 & 2.80 & 2.00 & 2.40 \\
 & fr & 50.20 & 50.80 & 50.20 & 50.40 & 25.19 & 21.66 & 21.66 & 22.84 & 2.80 & 2.80 & 1.20 & 2.27 \\
 & zh & 50.00 & 51.20 & 46.80 & 49.33 & 20.43 & 21.51 & 22.12 & 21.35 & 2.40 & 1.20 & 2.80 & 2.13 \\
 \midrule
\multirow{5}{*}{Pythia-1.4B} & Zero-shot & 50.00 & 50.00 & 50.00 & 50.00 & 20.28 & 20.89 & 20.28 & 20.48 & 2.00 & 2.00 & 1.60 & 1.87 \\
 & Average & 49.73 & 50.73 & 49.53 & 50.00 & 22.63 & 21.30 & 21.61 & 21.85 & 6.27 & 4.00 & 4.67 & 4.98 \\
 & en & 49.60 & 51.60 & 50.20 & 50.47 & 21.20 & 20.12 & 19.97 & 20.43 & 8.40 & 3.20 & 3.20 & 4.93 \\
 & fr & 49.80 & 51.00 & 49.80 & 50.20 & 25.04 & 19.51 & 23.50 & 22.68 & 6.80 & 8.00 & 1.60 & 5.47 \\
 & zh & 49.80 & 49.60 & 48.60 & 49.33 & 21.66 & 24.27 & 21.35 & 22.43 & 3.60 & 0.80 & 9.20 & 4.53 \\
 \midrule
\multirow{5}{*}{Pythia-6.9B} & Zero-shot & 50.00 & 50.40 & 50.00 & 50.13 & 21.97 & 22.27 & 20.28 & 21.51 & 4.80 & 3.20 & 2.00 & 3.33 \\
 & Average & 50.33 & 51.93 & 49.20 & 50.49 & 28.21 & 27.96 & 26.01 & 27.39 & 10.27 & 8.67 & 7.07 & 8.67 \\
 & en & 50.80 & 53.40 & 50.00 & 51.40 & 32.10 & 30.26 & 27.96 & 30.11 & 12.80 & 10.00 & 6.00 & 9.60 \\
 & fr & 50.00 & 51.00 & 49.20 & 50.07 & 25.65 & 27.19 & 24.42 & 25.76 & 10.80 & 11.60 & 4.40 & 8.93 \\
 & zh & 50.20 & 51.40 & 48.40 & 50.00 & 26.88 & 26.42 & 25.65 & 26.32 & 7.20 & 4.40 & 10.80 & 7.47 \\
 \midrule
\multirow{5}{*}{LLaMA-6.7B} & Zero-shot & 54.40 & 51.00 & 52.00 & 52.47 & 21.97 & 24.88 & 22.27 & 23.04 & 4.00 & 3.20 & 3.20 & 3.47 \\
 & Average & 72.00 & 63.67 & 54.07 & 63.24 & 33.59 & 32.10 & 29.03 & 31.58 & 22.93 & 18.00 & 10.13 & 17.02 \\
 & en & 85.60 & 71.40 & 59.80 & 72.27 & 37.63 & 33.79 & 31.34 & 34.25 & 27.20 & 18.00 & 7.20 & 17.47 \\
 & fr & 72.20 & 65.40 & 51.00 & 62.87 & 37.79 & 36.41 & 33.03 & 35.74 & 24.40 & 21.60 & 7.20 & 17.73 \\
 & zh & 58.20 & 54.20 & 51.40 & 54.60 & 25.35 & 26.11 & 22.73 & 24.73 & 17.20 & 14.40 & 16.00 & 15.87 \\
 \midrule
\multirow{5}{*}{LLaMA-13.0B} & Zero-shot & 62.20 & 52.20 & 50.60 & 55.00 & 25.35 & 26.42 & 20.28 & 24.01 & 5.20 & 3.20 & 3.20 & 3.87 \\
 & Average & 85.40 & 76.33 & 61.40 & 74.38 & 38.91 & 37.22 & 34.66 & 36.93 & 30.93 & 25.73 & 16.93 & 24.53 \\
 & en & 89.20 & 76.80 & 63.80 & 76.60 & 39.78 & 37.63 & 33.03 & 36.82 & 34.40 & 27.60 & 13.20 & 25.07 \\
 & fr & 83.00 & 75.20 & 58.80 & 72.33 & 40.40 & 36.87 & 35.18 & 37.48 & 31.60 & 30.40 & 15.20 & 25.73 \\
 & zh & 84.00 & 77.00 & 61.60 & 74.20 & 36.56 & 37.17 & 35.79 & 36.51 & 26.80 & 19.20 & 22.40 & 22.80 \\
 \midrule
\multirow{5}{*}{LLaMA-32.5B} & Zero-shot & 50.00 & 50.00 & 50.00 & 50.00 & 20.58 & 27.19 & 21.20 & 22.99 & 15.20 & 10.00 & 3.20 & 9.47 \\
 & Average & 95.13 & 91.27 & 76.07 & 87.49 & 49.51 & 46.80 & 43.57 & 46.63 & 46.80 & 43.60 & 29.87 & 40.09 \\
 & en & 95.40 & 90.00 & 73.80 & 86.40 & 50.54 & 45.93 & 43.32 & 46.59 & 46.80 & 46.40 & 26.00 & 39.73 \\
 & fr & 95.40 & 93.00 & 77.20 & 88.53 & 51.15 & 47.93 & 41.32 & 46.80 & 51.20 & 45.60 & 28.40 & 41.73 \\
 & zh & 94.60 & 90.80 & 77.20 & 87.53 & 46.85 & 46.54 & 46.08 & 46.49 & 42.40 & 38.80 & 35.20 & 38.80 \\
 \bottomrule
\end{tabular}}
\end{table}

The detailed results for the three BLOOM, Pythia, and LLaMA models across 15 languages on the XNLI dataset are shown in Table~\ref{tab:bloom-xnli}, Table~\ref{tab:pythia-xnli}, and Table~\ref{tab:llama-xnli}, respectively. The results on other three datasets (i.e., GSM8K, XCOPA and LogiQA) are listed in Table~\ref{tab:others}.

\section{More Figures}
\begin{figure}[ht]
\centering
\includegraphics[width=5.5in]{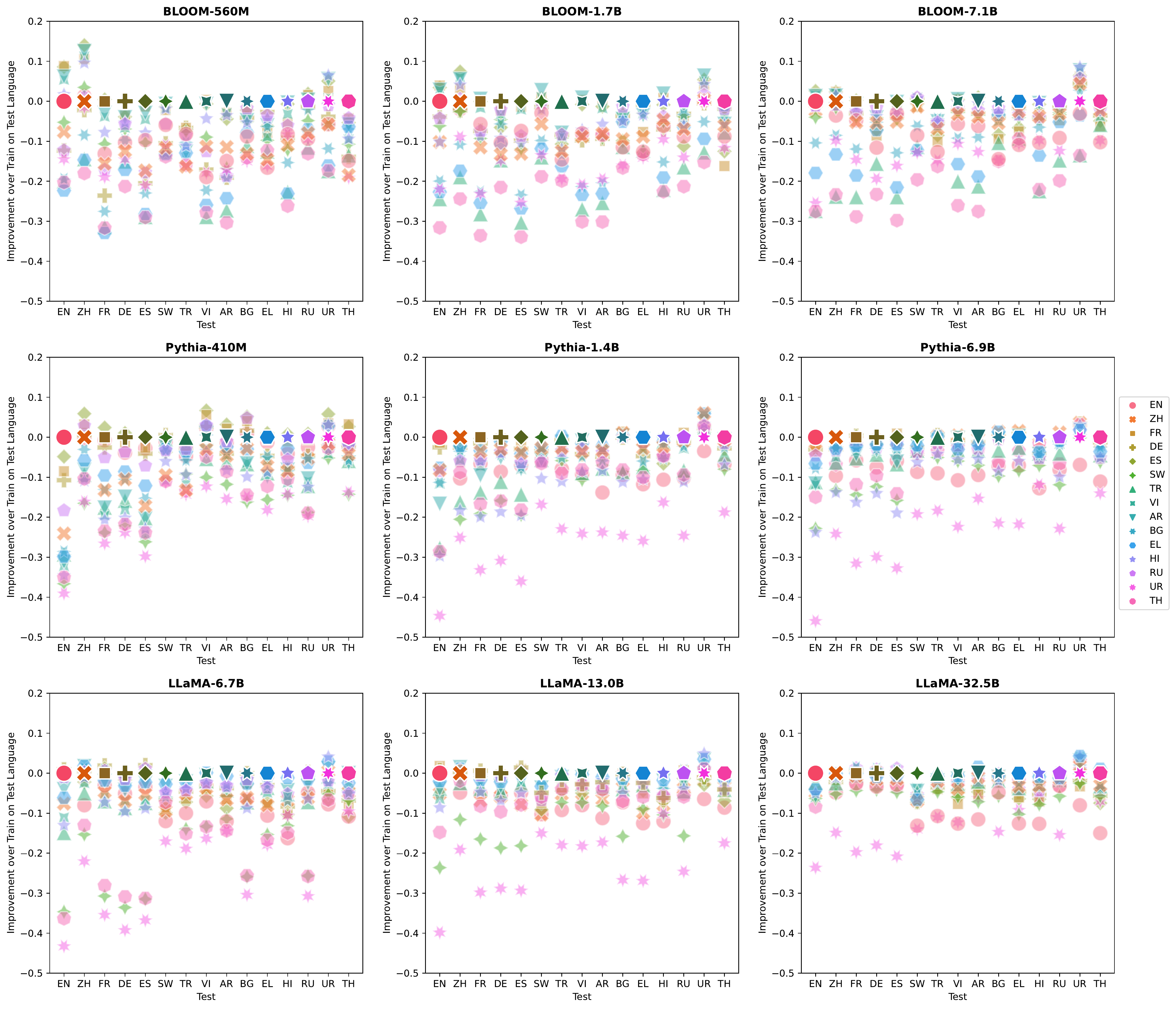}
\caption{Accuracy gain of BLOOMs and LLaMAs on test languages by subtracting the performance of models trained on each test language from those trained on other languages.
}
\label{fig:on-target}
\end{figure}

\begin{figure}[ht]
\centering
\includegraphics[width=5.5in]{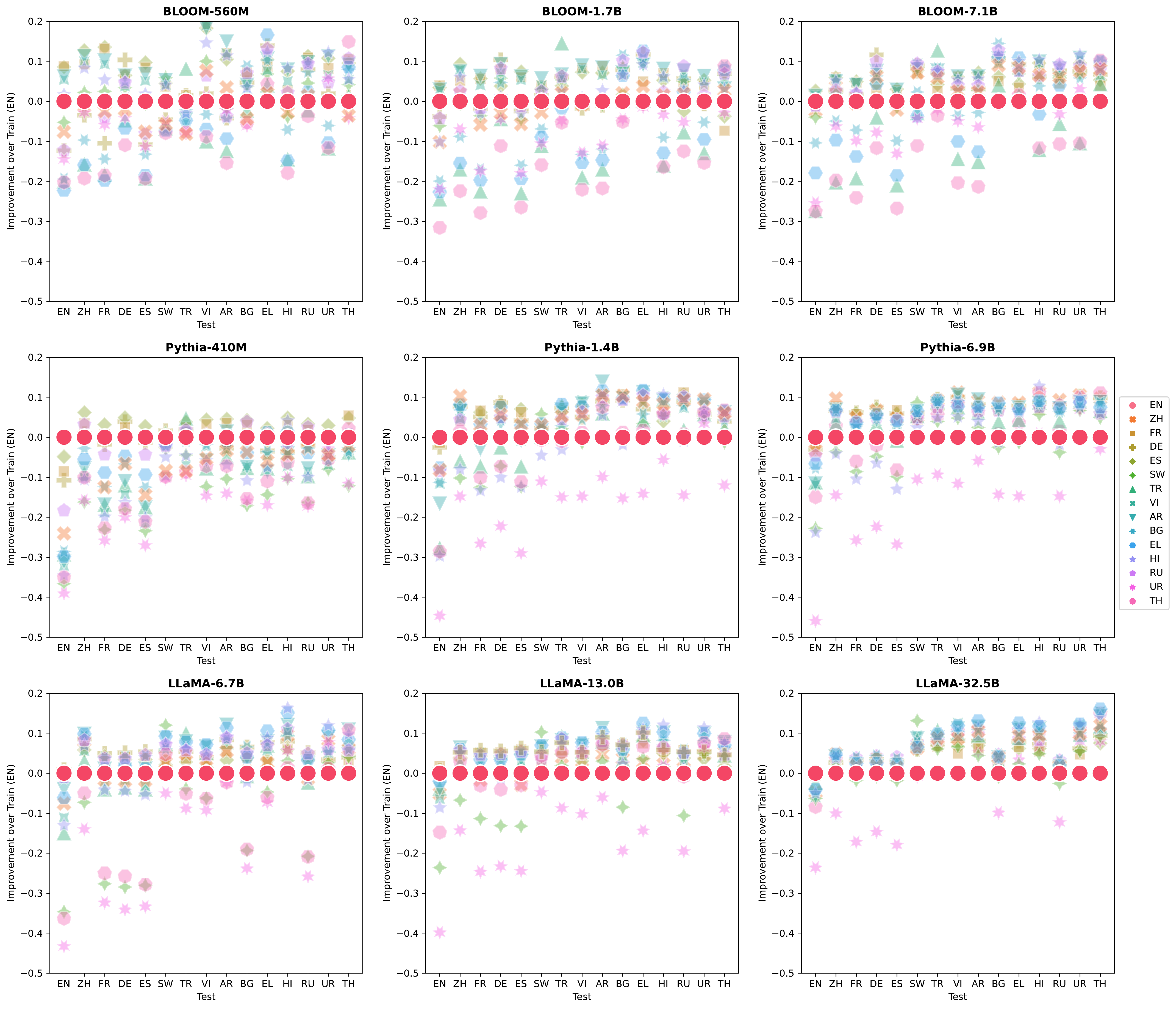}
\caption{Accuracy gain of BLOOMs and LLaMAs on test languages by subtracting the performance of models trained on English from those trained on other languages. 
}
\label{fig:on-english}
\end{figure}

We show the accuracy gain of BLOOMs and LLaMAs on test languages by subtracting the performance of models trained on each test language from those trained on other languages in Figure~\ref{fig:on-target}. This figure is complementary to Figure 3 which only shows the results for BLOOMs and LLaMAs in the paper.
Similarly, we also show the accuracy gain by subtracting the performance of models trained on English from those trained on other languages in Figure~\ref{fig:on-english}. This figure corresponds to the average number of superior training languages compared with English in Figure 2 of the paper, and shows specifically which languages are better used for training given a test language. 
 

\end{document}